\documentclass{article} 
\usepackage{iclr2026_conference,times}

\usepackage{amsmath,amsfonts,bm}

\def\eqref#1{equation~\ref{#1}}

\def\1{\bm{1}}

\DeclareMathAlphabet{\mathsfit}{\encodingdefault}{\sfdefault}{m}{sl}
\SetMathAlphabet{\mathsfit}{bold}{\encodingdefault}{\sfdefault}{bx}{n}

\definecolor{tablegray}{gray}{0.9}
\usepackage{hyperref}
\usepackage{url}
\usepackage{tcolorbox}
\usepackage{tabularx}
\usepackage{amssymb}
\usepackage{algorithm}
\usepackage{algorithmic}
\usepackage{amsmath}
\usepackage{booktabs} 
\usepackage[utf8]{inputenc}
\usepackage[svgnames]{xcolor} 
\usepackage{tcolorbox}       
\definecolor{mypink}{RGB}{255, 228, 225}
\usepackage{booktabs} 
\usepackage{subcaption} 
\usepackage{xcolor}
\definecolor{VeryLightGray}{gray}{0.9}
\usepackage{amssymb}  
\usepackage{graphicx} 
\usepackage{multirow} 
\usepackage{makecell} 
\usepackage[table]{xcolor} 
\usepackage{paralist}
\title{Can Vision–Language Models Assess Graphic Design Aesthetics? A Benchmark, Evaluation, and Dataset Perspective}

\author{%
     \bf Arctanx An\textsuperscript{1,2}\thanks{Work was done during an internship at Microsoft Research Asia}
  ~~ \bf{Shizhao Sun}\textsuperscript{2}\thanks{Project Leader}
  ~~ \bf Danqing Huang\textsuperscript{3}
  ~~ \bf Mingxi Cheng\textsuperscript{3} \\ 
  ~~ \bf Yan Gao\textsuperscript{4} 
  ~~ \bf Ji Li \textsuperscript{3}
  ~~ \bf Yu Qiao\textsuperscript{4}
  ~~ \bf Jiang Bian \textsuperscript{2}
  \\
  \textsuperscript{1}Peking University ~~~
  \textsuperscript{2}Microsoft Research Asia ~~~
  \textsuperscript{3}Microsoft ~~~
  \textsuperscript{4}Central South University
}

\iclrfinalcopy 

\begin{document}

\maketitle

\begin{abstract}
Assessing the aesthetic quality of graphic design is central to visual communication, yet remains underexplored in vision–language models (VLMs). 
We investigate whether VLMs can evaluate design aesthetics in ways comparable to humans. 
Prior work faces three key limitations: benchmarks restricted to narrow principles and coarse evaluation protocols, a lack of systematic VLM comparisons, and limited training data for model improvement. 
In this work, we introduce AesEval-Bench, a comprehensive benchmark spanning four dimensions, twelve indicators, and three fully quantifiable tasks: aesthetic judgment, region selection, and precise localization. 
Then, we systematically evaluate proprietary, open-source, and reasoning-augmented VLMs, revealing clear performance gaps against the nuanced demands of aesthetic assessment. 
Moreover, we construct a training dataset to fine-tune VLMs for this domain, leveraging human-guided VLM labeling to produce task labels at scale and indicator-grounded reasoning to tie abstract indicators to concrete design regions.
Together, our work establishes the first systematic framework for aesthetic quality assessment in graphic design. 
Our code and dataset will be released at: \href{https://github.com/arctanxarc/AesEval-Bench}{https://github.com/arctanxarc/AesEval-Bench}. 

\end{abstract}

\section{Introduction}
The rapid development of \emph{vision-language models (VLMs)}~\citep{yang2025qwen3, li2024llava, hurst2024gpt} has opened new opportunities for understanding and acting upon multimodal information. 
While they have already achieved remarkable progress in traditional vision tasks such as image captioning~\citep{lin2024draw,  li2025chemvlm} and visual question answering~\citep{an2024mc, zhang2024document, lin2025perceive}, VLMs are increasingly expected to contribute to new applications. 
Among these, \emph{graphic design}---which integrates textual and visual elements to convey information across advertising, branding, and digital media---represents a promising direction given its broad societal and practical impact. 
Crucially, the success of graphic design critically depends on its \emph{aesthetic quality}, shaped by principles such as balance, contrast, and hierarchy.

In this work, we aim to explore a central question: \emph{can VLMs understand and evaluate aesthetic quality of graphic design in a manner comparable to humans?} 
This question is of significant importance for at least three reasons. 
First, it can assist human designers by identifying where a design falls short and explaining why, thereby enabling more effective improvement. 
Second, for generative AI systems, it provides the basis for automatic feedback loops that can guide iterative refinement without extensive human intervention. 
Third, it suggests an opportunity to extend VLMs beyond factual recognition toward aesthetic evaluation.

Despite its importance, research along this direction remains limited (see Table~\ref{tab:benchmark_comparison_centered}). 
First, \emph{benchmarks are inadequate}. 
Those~\citep{zhou2024uniaa, huang2024aesbench} developed for natural photos often ignore design-specific factors such as typography, while early ones for graphic design typically cover only a narrow subset of design principles~\citep{lin2023designbench, lin2024designprobe, jiang2025multimodal}. 
Moreover, existing evaluation protocols are limited.
Scoring-based methods fail to indicate where poor aesthetics occur~\citep{haraguchi2024can}, while description-based ones provide qualitative feedback that is difficult to quantify~\citep{jung2025ui}. 
Second, \emph{comparisons between VLMs are missing}.
There has been no systematic evaluation across different VLMs, whether open-source or closed-source, in the context of design aesthetics. 
Third, \emph{training datasets are lacking}, leaving the question of how to further improve VLM performance in this domain underexplored.

As a first step toward addressing these limitations, we introduce \emph{AesEval-Bench}, a new benchmark for evaluating the aesthetic quality of graphic designs (see Figure~\ref{fig:benchmark overview}).
Drawing on prior literature~\citep{li2009aesthetic, wangwiwattana2024design}, we identify four critical \underline{\emph{dimensions}}---typography, layout, color, and graphics---that together comprehensively capture design aesthetics. 
These dimensions reflect the major factors that humans consistently emphasize when assessing visual appeal. 
To provide finer granularity, we further define twelve \underline{\emph{indicators}} that specify concrete aspects within each dimension, such as hierarchy and legibility under typography.
For each indicators, we then design three challenging \underline{\emph{tasks}}:
1) \emph{aesthetic judgment}, which asks models to decide whether a design is aesthetically pleasing (yes/no), providing a straightforward measure of overall perception; 
2) \emph{region selection}, which requires models to choose from candidate regions where unpleasing elements appear, testing their ability to pinpoint problematic areas beyond a global judgment; 
3) \emph{precise localization}, which challenges models to predict the exact bounding box (bbox) coordinates of unpleasing areas, offering the most detailed diagnosis and reflecting a deeper understanding of aesthetics.
Unlike prior benchmarks~\citep{}, AesEval-Bench not only covers a broad range of aesthetic factors through its dimensions and indicators, but also defines evaluation tasks that are fully quantifiable via choice or bbox prediction formats, enabling systematic and reproducible assessment of design aesthetics.

With AesEval-Bench, we systematically evaluate VLMs on their ability to assess the aesthetic quality of graphic designs (see Table~\ref{tab: compare}).
For each design, models perform three tasks across twelve indicators. 
We use human annotation as ground truth, and measure performance by accuracy (for aesthetic judgment and region selection) and bbox IoU (for precise localization).
Overall, our results reveal clear gaps between current state-of-the-art VLMs and the nuanced demands of aesthetic quality assessment.
Specifically, proprietary VLMs (e.g., GPT series~\citep{achiam2023gpt}) outperform open-source ones (e.g., Qwen-VL~\citep{wang2024qwen2}, Intern-VL~\citep{zhu2025internvl3}, LLaVA~\citep{liu2023visual}). 
Among open-source models, larger variants (32B, 72B) generally achieve better performance than smaller ones (7B).
Surprisingly, reasoning-augmented VLMs (e.g., GPT-o1~\citep{jaech2024openai}, GPT-o3, Gemini-2.5-Pro~\citep{comanici2025gemini}) offer no clear advantage over their non-reasoning counterparts.
Together, these findings expose the limitations of existing VLMs and underscore the need for domain-specific training tailored to aesthetic quality assessment.

Building on these findings, we move beyond evaluation and turn to training VLMs for aesthetic quality assessment. 
To this end, we construct a training dataset consisting of three components: the \emph{task} (what the model is asked to do), the \emph{task label} (the expected answer), and the \emph{reasoning path} (the explanation leading to the answer) (see Figure~\ref{fig:training}). 
We treat the reasoning path as essential, since generic reasoning has shown little benefit.
However, constructing such data poses two key challenges: producing task labels at scale is costly, and generating reasoning paths that genuinely improve performance requires new approaches. 
To address these, we introduce two solutions.
First, \emph{human-guided VLM labeling}, where a small set of human annotations serve as in-context examples to instruct powerful VLMs in producing task labels. 
This approach maintains alignment with human understanding while reducing manual annotation costs. 
Second, \emph{indicator-grounded reasoning}, where abstract indicators (e.g., hierarchy or layering) are explicitly tied to concrete regions in the design. 
Each reasoning path consists of bounding-box coordinates linked to the indicator and textual explanations for their relevance, providing fine-grained and interpretable supervision.
We fine-tune VLMs with the task as input and both the reasoning path and the task label as supervision, and evaluate on AesEval-Bench. 
The results show consistent performance gains across all tasks and indicators (e.g., 5.97\%, 2.70\% and 17.17\%), demonstrating that human-guided VLM labeling yields reliable labels and indicator-grounded reasoning supplies effective supervision.

To sum up, our contributions are as follows:
\begin{compactitem}
\item We introduce AesEval-Bench, a comprehensive benchmark for assessing aesthetic quality of graphic designs spanning four dimensions, twelve indicators and three quantifiable tasks.
\item We systematically evaluate proprietary, open-source, and reasoning-augmented VLMs, revealing clear performance gaps in aesthetic quality assessment. 
\item We construct a training dataset to fine-tune VLMs for this domain. Our approach introduces human-guided VLM labeling to produce task labels at scale and indicator-grounded reasoning to tie abstract indicators to concrete design regions.
Experiments on AesEval-Bench show that this dataset consistently improves performance across all tasks.  
\end{compactitem}

\begin{table*}[t]
\centering
\caption{
A comparison of AesEval-Bench with existing benchmarks for both image aesthetics and design aesthetics. 
We highlight key differences in their scale, task formats, source data, covered design dimensions, and the inclusion of reasoning paths.
}
\label{tab:benchmark_comparison_centered}
\renewcommand{\arraystretch}{1.5}
\resizebox{\textwidth}{!}{%
\begin{tabular}{cccccccccccc}
\toprule
\multirow{2}{*}{\textbf{Benchmark}} & \multirow{2}{*}{\textbf{\#Data}} & \multirow{2}{*}{\makecell{\textbf{Task}\\\textbf{Format}}} & \multirow{2}{*}{\textbf{Source}} & \multirow{2}{*}{\textbf{Source Type}} & \multicolumn{4}{c}{\textbf{Dimension}} & \multirow{2}{*}{\textbf{Training Set}} & \multirow{2}{*}{\textbf{Reasoning Path}} & \multirow{2}{*}{\textbf{Open-source}} \\
\cmidrule(lr){6-9}
& & & & & \textbf{Font} & \textbf{Layout} & \textbf{Graphics} & \textbf{Color} & & & \\
\midrule
\rowcolor{black!5}
\multicolumn{12}{c}{\textit{Image Aesthetics Benchmark}} \\
\midrule
\makecell{AesBench \\ ~\citep{huang2024aesbench}} & $\sim$10k & \makecell{Free-form} & \makecell{Photographic \\ Image} & Image-only & $\times$ & $\times$ & $\checkmark$ & $\checkmark$ & $\times$ & $\times$ & $\checkmark$ \\
\makecell{UNIAA-Bench \\ ~\citep{zhou2024uniaa}} & $\sim$6k & \makecell{Free-form} & \makecell{Photographic \\ Image} & Image-only & $\times$ & $\checkmark$ & $\times$ & $\checkmark$ & $\times$ & $\times$ & $\checkmark$ \\
\makecell{FineArtBench \\ ~\citep{jiang2025multimodal}} & - & \makecell{Choice \\ Free-form} & \makecell{Art Work + \\ Photographic Image} & Image-only & $\times$ & $\times$ & $\times$ & $\checkmark$ & $\times$ & $\checkmark$ & $\times$ \\
\midrule
\rowcolor{black!5}
\multicolumn{12}{c}{\textit{Design Aesthetics Benchmark}} \\
\midrule
\makecell{DesignBench \\ ~\citep{lin2023designbench}} & - & \makecell{Choice \\ Free-form} & \makecell{Graphic Design} & Image+Json & $\checkmark$ & $\checkmark$ & $\times$ & $\checkmark$ & $\times$ & $\times$ & $\checkmark$ \\
\makecell{DesignProbe \\ ~\citep{lin2024designprobe}} & $\sim$1.6k & \makecell{Choice} & \makecell{Graphic  Design} & Image+Json & $\checkmark$ & $\checkmark$ & $\times$ & $\checkmark$ & $\times$ & $\times$ & $\times$ \\
\makecell{GPT-Eval Bench \\ ~\citep{haraguchi2024can}} & $\sim$2k & \makecell{Scoring} & \makecell{Graphic  Design} & Image+Json & $\times$ & $\checkmark$ & $\times$ & $\times$ & $\times$ & $\times$ & $\times$ \\
\makecell{UI-Bench \\ ~\citep{jung2025ui}} & $\sim$3k & \makecell{Choice \\ Description} & \makecell{UI Design} & Image-only & $\checkmark$ & $\checkmark$ & $\times$ & $\checkmark$ & $\times$ & $\times$ & $\times$ \\
\makecell{UICrit \\ ~\citep{jung2025ui}} & $\sim$3k & \makecell{Free-form \\ Bbox regression} & \makecell{UI Design} & Image-only & $\checkmark$ & $\checkmark$ & $\times$ & $\checkmark$ & $\times$ & $\times$ & $\checkmark$ \\
AesEval-Bench (Ours) & $\sim$4.5k & \makecell{Choice \\ Bbox Regression} & \makecell{Graphic  Design} & Image+Json & $\checkmark$ & $\checkmark$ & $\checkmark$ & $\checkmark$ & $\checkmark$ & $\checkmark$ & $\checkmark$ \\
\bottomrule
\end{tabular}%
}
\label{tab: compare}
\end{table*}

\section{Related Works}

\textbf{Aesthetic Quality Assessment.}
Aesthetic quality assessment~\citep{deng2017image} aims to automatically evaluate visual appeal, serving as a computational proxy for human judgment. 
Within this area, two major lines of research have emerged (see Table~\ref{tab:benchmark_comparison_centered}). 
\emph{Image aesthetics assessment}~\citep{huang2024aesbench, zhou2024uniaa, jiang2025multimodal} focuses on photographic images, where quality is determined by factors such as color harmony, lighting, and subject placement. 
\emph{Design aesthetics assessment}~\citep{lin2023designbench, lin2024designprobe, haraguchi2024can, jung2025ui} targets graphic designs such as posters, advertisements, or user interfaces, which depend on design-related factors including typography, hierarchy, and alignment.
Our work falls within design aesthetics assessment.

Despite its importance, design aesthetics assessment remains underexplored. 
Existing benchmarks capture only a narrow subset of design dimensions.
For instance, \citep{ lin2024designprobe} omits graphics-related factors, while \citep{haraguchi2024can} ignores both fonts and graphics. 
Furthermore, their task formulations lack rigor. 
Some adopt free-form question answering, which is difficult to quantify~\citep{lin2023designbench}, while others provide only holistic scores without identifying problematic regions, limiting interpretability and actionability~\citep{jung2025ui}. 
Our work introduces a benchmark that comprehensively covers design-related aesthetic factors across font, layout, graphics and color, defining well-structured and quantifiable tasks using choice and bbox prediction formats.

\textbf{Vision-Language Models.}
Vision-Language Models (VLMs)~\citep{wang2024qwen2, comanici2025gemini, li2024llava, zhang2025latent} have achieved remarkable performance on tasks such as image captioning~\citep{luo2025drvhierarchicalperceptiontemporalcognitionframework, li2024survey, zhangmme} and visual question answering~\citep{an2024mc, lin2024draw, luo2025dr}. 
Yet, their ability to assess the aesthetic quality of graphic designs remains largely unexplored. 
Prior work typically evaluates only one or two VLMs (e.g., \citep{haraguchi2024can} studies GPT). 
We provide a systematic comparison across a broad set of VLMs, including proprietary, open-source, and reasoning-augmented models.

Recently, increasing attention has been devoted to the reasoning capabilities of VLMs~\citep{li2025imagine, zhang2025scaling}.
For example, \citep{sarch2025grounded} employs tree-based search to improve reasoning chains, \citep{li2025imagine} visualizes reasoning trajectories for transparency, and \citep{sun2024visual, shao2024visual, cao2025ground, wu2025grounded} explores grounded visual reasoning by jointly generating bounding boxes and textual explanations.
In our work, we observe that generic reasoning in current VLMs provides limited benefit for assessing design aesthetics.
To address this, we construct a training dataset with reasoning paths that explicitly link abstract design indicators to concrete regions of the design.
Unlike grounded visual reasoning, which localizes semantically salient entities (e.g., a ``dog'' or ``chair''), our regions are indicator-centric, capturing higher-level concepts such as hierarchy, alignment, and spacing that directly embody design principles.



\section{Benchmark Construction}

\subsection{Overview}
\label{subsec:benchmark_overview}
\textbf{AesEval-Bench} formulates design aesthetics assessment as a question–answering task. 
The input contains a \emph{task} description and a \emph{design image}, optionally accompanied by metadata such as layout, font, or color information in JSON format. 
The output is the \emph{answer} corresponding to the task.

\begin{figure}[t]
    \centering
    \includegraphics[width=1\textwidth]{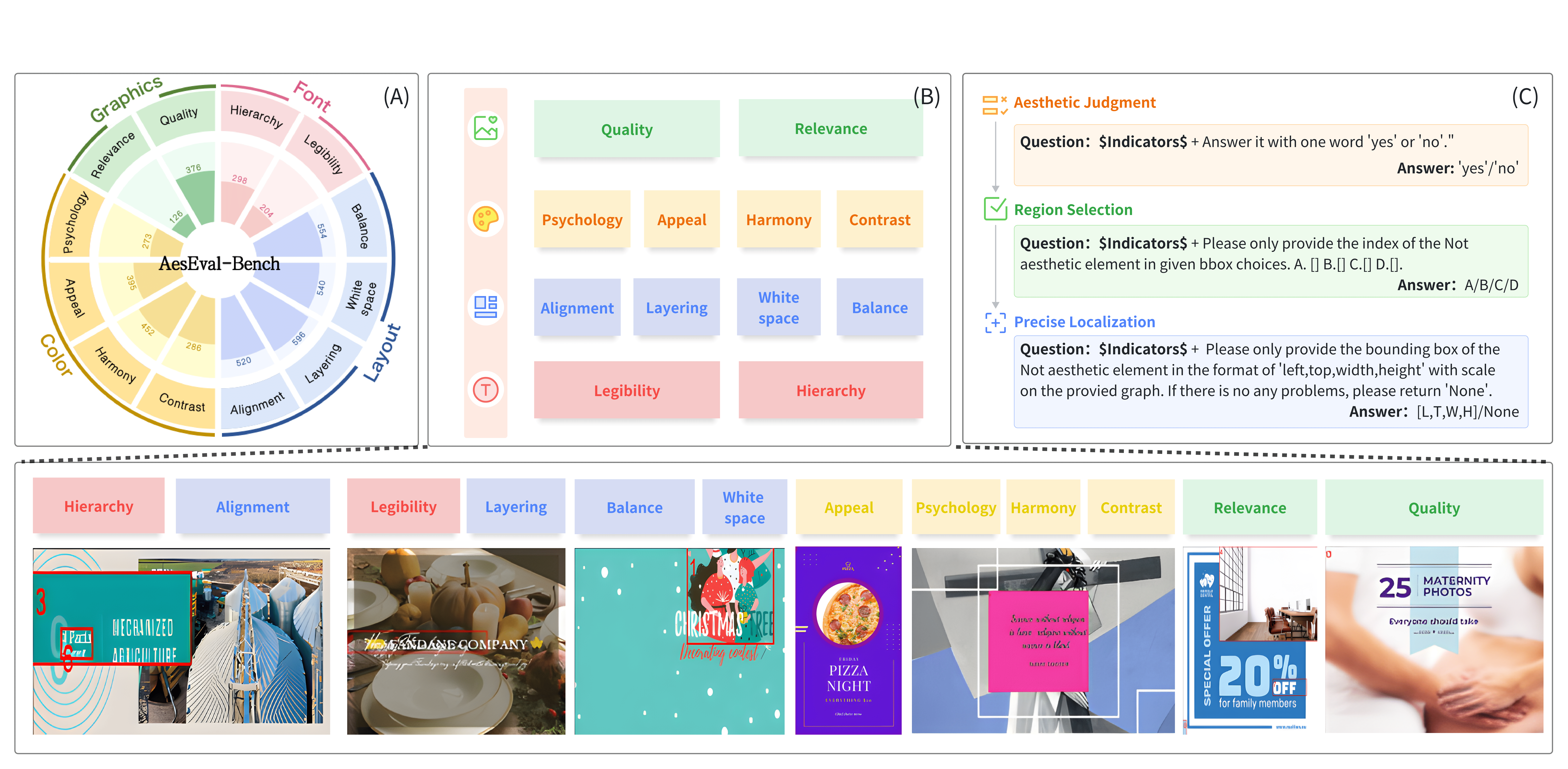}
    \caption{
    Overview of AesEval-Bench.
\textbf{(A)} The four dimensions and twelve indicators considered in the benchmark. Numbers inside the circles indicate how many designs are labeled as flawed for each indicator.
\textbf{(B)} Example designs illustrating the indicators, with regions exhibiting aesthetic issues highlighted by red boxes. Detailed textual explanations of all indicators are provided in the Appendix.
\textbf{(C)} The three tasks, along with example questions and their expected answers.
    }
    \label{fig:benchmark overview}
    \vspace{-6mm}
\end{figure}

To capture different aspects of design aesthetics assessment, we introduce three task types (Figure~\ref{fig:benchmark overview}(C)):
1) \emph{aesthetic judgment} asks whether a design is aesthetically pleasing (yes/no), providing a measure of overall perception.
2) \emph{region selection} requires choosing from candidate regions where aesthetic issues appear, testing the ability to localize problematic areas beyond a global judgment.
3) \emph{precise localization} requires predicting the exact bounding box coordinates of problematic regions, offering a fine-grained diagnosis.
Each task is accompanied by the explanation of an indicator---the key factor humans consistently emphasize when evaluating visual appeal (e.g., hierarchy, layering, contrast). 
We consider twelve indicators (Figure~\ref{fig:benchmark overview}(B)), grouped into four dimensions (Figure~\ref{fig:benchmark overview}(A)).

For design images, we sample 1200 designs from the test split of Crello dataset~\citep{yamaguchi2021canvasvae}, which contains professional graphic designs with both the design image and its metadata. 
The expected answers differ across tasks. 
For aesthetic judgment, the answer is yes or no. For region selection, it is the index of one region among four candidates. 
For precise localization, it is the bounding box coordinates of the identified region or None if the design has no aesthetic issues.

Overall, AesEval-Bench comprises 4500 base question–answer pairs (three tasks across 1500 designs), each further instantiated across twelve indicators to enable fine-grained evaluation.

\subsection{Curation Pipeline}
\label{subsec:benchmark_pipeline}

\begin{figure}[h!]
    \centering
    \includegraphics[width=1\textwidth]{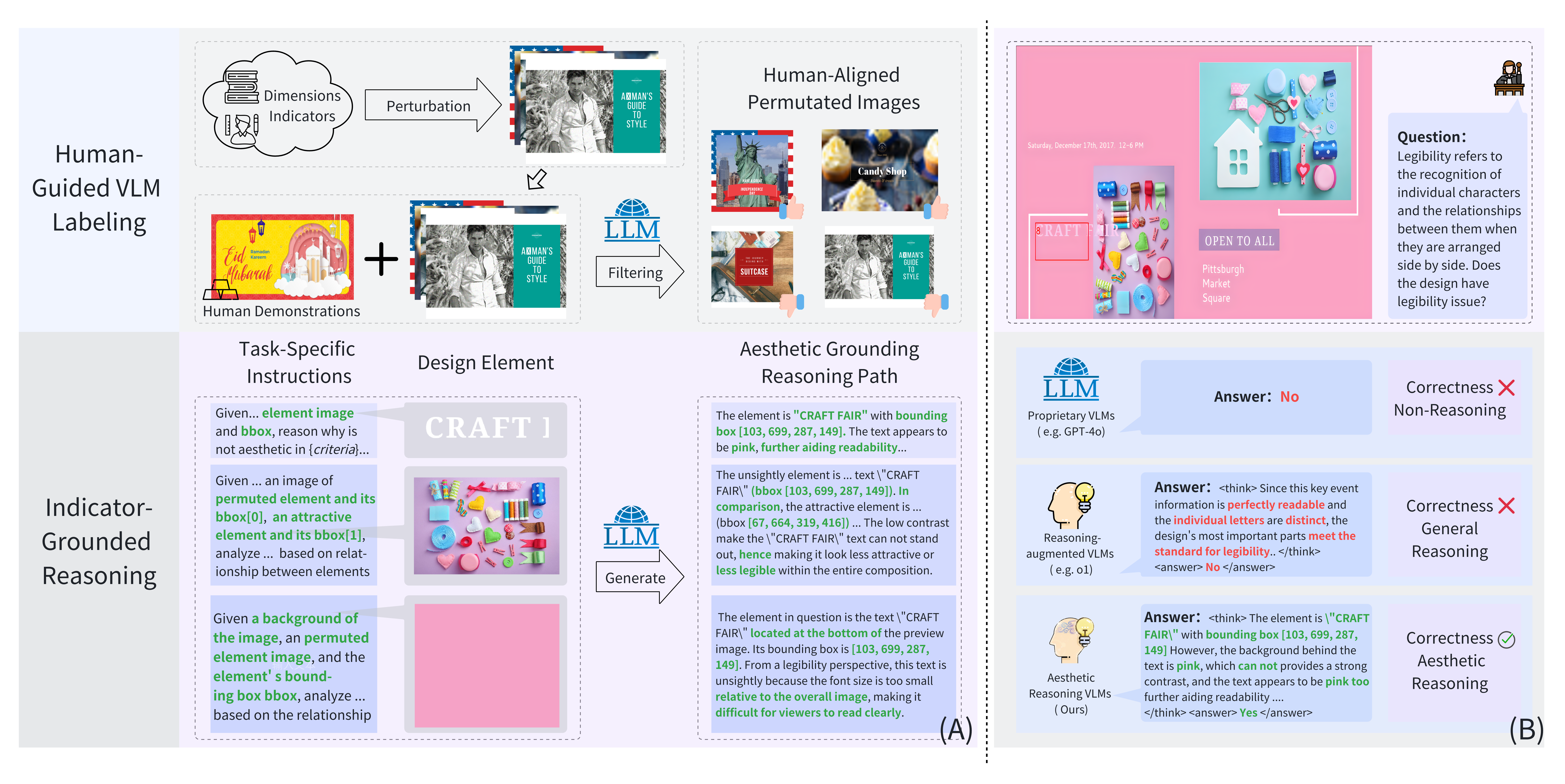}
    \vspace{-4mm}
    \caption{
 \textbf{(A)} Illustration of two key steps in training data construction. Human-guided VLM labeling enables scalable determination of whether designs exhibit aesthetic issues. Indicator-grounded reasoning generates reasoning paths that explicitly link abstract indicators to concrete design regions (represented as bbox coordinates).
\textbf{(B)} Example highlighting the difference between non-reasoning models, generic reasoning models, and our indicator-grounded reasoning model.
    }
    \label{fig:training} 
    \vspace{-6mm}
\end{figure}

\textbf{Establishing Dimensions and Indicators.}
Aesthetic quality in graphic design is inherently multidimensional.
To define a rigorous benchmark, we first conducted a comprehensive literature review of classical and contemporary design principles~\citep{mccormack2020understanding, lou2022designeva, lu2020computational}.
We then consulted professional designers to refine this taxonomy, ensuring alignment with both theoretical foundations and practical expertise.
This process yielded four core dimensions---layout, font, graphics, and color (Figure~\ref{fig:benchmark overview}(A))---each further specified by concrete indicators consistently emphasized in human aesthetic judgment.
In total, we distilled twelve indicators that together capture the essential factors of design aesthetics (Figure~\ref{fig:benchmark overview}(B)).

\textbf{Constructing Potentially Flawed Designs.}
\label{sec: flaw}
As introduced in Section~\ref{subsec:benchmark_overview}, the design images in AesEval-Bench are sourced from the Crello dataset, which contains professional-quality graphic designs.
To effectively evaluate design aesthetics, the benchmark must include not only well-designed but also less appealing examples.
We therefore repurpose Crello by introducing controlled perturbations, such as repositioning elements, altering font choices, or adjusting colors.
These perturbations may either degrade the visual quality or leave it largely intact.
For example, slightly enlarging a heading might preserve hierarchy, whereas shifting it left could disrupt balance.
Each base design undergoes one to three random perturbations, generating a spectrum of variations that range from aesthetically unchanged to noticeably flawed, while still appearing realistic.
Since Crello~\citep{yamaguchi2021canvasvae} provides element-level metadata in JSON format along with separate design layers, these perturbations can be applied directly at the JSON level and rendered into new design images by recombining the modified metadata with the corresponding layers.

\textbf{Human-in-the-Loop Aesthetic Review.}
We engage human annotators to verify whether the perturbed designs truly exhibit aesthetic issues.
Before annotation, all annotators receive a tutorial that includes examples of both well-designed and flawed cases, along with detailed explanations of the underlying reasons.
During the review, each annotator is shown a design image together with a description of the focal indicator and asked to determine whether the design contains the corresponding flaw (yes or no).
For each design, we derive the final label by applying majority voting across multiple annotators to ensure consensus.

\textbf{Generating Question-Answer Pairs.}
With metadata in JSON format, records of applied perturbations and human annotations, we can systematically construct answers corresponding to each task.
For aesthetic judgment, the rule is straightforward: if human annotators label a design as good, the ground-truth answer is no (i.e., no aesthetic issue); otherwise, it is yes.
For region selection, if a design is labeled as good, the four answer choices consist of three randomly sampled bboxes from the metadata and a None option, with the ground-truth answer being None. 
If the design is labeled as flawed, the four choices include the bbox of the perturbed element, two randomly sampled bboxes from the metadata, and None, with the ground-truth answer set to the bbox of the perturbed element.
For precise localization, if a design is labeled as good, the ground-truth answer is None; otherwise, it corresponds to the exact bbox of the perturbed element.

\subsection{Evaluation Protocols}
\label{subsec:eval_protocol}
For aesthetic judgment and region selection, both formulated as choice problems, we adopt accuracy as the metric, measuring the exact match between model predictions and the ground truth.
For precise localization, the task combines two components: a choice problem (predicting None when no aesthetic issue exists) and a bounding box regression problem (predicting the exact bbox when an issue is present). 
Accordingly, we use accuracy for cases where the ground truth is None, and intersection over union (IoU)---which quantifies the overlap between the predicted and ground-truth bboxes---for cases where a bbox is required. 


\section{Training Data Construction}
\label{sec:train_data}

Evaluation on popular VLMs  reveals clear gap between the capabiliteis of current state-of-the-arts VLMs and the nuanced requirements of aesthetic quality assessment.
Moreover, reasoning-augmented VLMs show no clear performance gains (see Section~\ref{sec: result}).

To this end, we construct a training dataset, named \textbf{AesEval-Train}, to fine-tune VLMs for this domain.
First, we adopt the same procedure as benchmark construction to \emph{construct potentially flawed designs} (see Section~\ref{sec: flaw}).
Next, since relying solely on human annotation to determine whether perturbed designs exhibit aesthetic issues is neither scalable nor cost-effective for training at large scale, we introduce \emph{human-guided VLM labeling}.
Then, we follow the benchmark construction to \emph{generate question–answer pairs} (see Section~\ref{sec: flaw}).
Finally, we introduce \emph{indicator-grounded reasoning} to generate domain-specific reasoning paths aimed at improving task performance.
In the following, we describe in detail the two steps that differ from benchmark construction. 

\textbf{Human-Guided VLM Labeling.} 
We leverage a small set of human annotations as demonstrations, together with the bbox coordinates of perturbed regions, as input to strong VLMs. 
The model is instructed to generate a binary label indicating whether the perturbed design exhibits an aesthetic issue (see Figure~\ref{fig:training}(A)). 
By incorporating human annotations, we preserve alignment with human judgment while substantially reducing manual annotation costs. 
Moreover, providing the perturbation region as a prior, which is unavailable in real-world scenarios, simplifies the labeling process and improves reliability. 
With these two sources of guidance, while the generated labels may not be perfectly accurate, they yield a training set of sufficient quality to enhance fine-tuning performance.


\textbf{Indicator-Grounded Reasoning.} 
As illustrated in Figure~\ref{fig:training}(B), generic reasoning often explains or analyzes a given indicator and task without grounding the discussion in relevant regions of the design. 
To address this limitation, we propose explicitly linking abstract indicators to concrete regions within the design. 
Specifically, we include both the bounding box (bbox) coordinates of relevant regions and textual explanations of their relevance to the indicator in the reasoning path. 

To obtain such reasoning paths, we instruct powerful VLMs (e.g., GPT in our experiments) by providing them with the bbox coordinates of the target regions and the corresponding design layers (Figure~\ref{fig:training}(A)). 
The model is required to output the provided coordinates alongside an explanation of how the region relates to the indicator, thereby ensuring that the reasoning path consistently contains the desired information.
We further adopt task-specific strategies to determine the regions of interest. 
For aesthetic judgment, we directly use the bbox of the perturbed regions. For region selection, we include both the perturbed and non-perturbed regions to strengthen the model’s ability to discriminate among candidate regions. 
For precise localization, we not only highlight the bbox of perturbed regions but also emphasize their relationship to the overall design, enabling the model to better localize problematic regions within the global design context.

\begin{table*}[t]
\centering
\caption{Evaluation on aesthetic judgment task. Overall acc is the average value of all indicators. The best and second-best results are highlighted in \textbf{bold} and \underline{underlined}, respectively.}
\label{tab:model_design_evaluation}
\resizebox{\textwidth}{!}{%
\begin{tabular}{c c cccc cccc cc cc}
\toprule
\multirow{2}{*}{\textbf{Model}} & \multirow{2}{*}{\textbf{Overall Acc}} & \multicolumn{4}{c}{\textbf{Layout}} & \multicolumn{4}{c}{\textbf{Color}} & \multicolumn{2}{c}{\textbf{Font}} & \multicolumn{2}{c}{\textbf{Graphics}} \\
\cmidrule(lr){3-6} \cmidrule(lr){7-10} \cmidrule(lr){11-12} \cmidrule(lr){13-14}
& & balance & layering & whitespace & alignment & harmony & contrast & appeal & psycholoy & legibility & hierarchy & quality & relevance \\
\midrule
\rowcolor{black!5}
\multicolumn{14}{c}{\textbf{Non-reasoning Models}} \\
\midrule
LLaVA-13B & 0.5636 & 0.6506 & 0.5063 & 0.6975 & 0.6759 & 0.4411 & 0.2409 & 0.2615 & 0.2851 & 0.7660 & 0.8260 & 0.7386 & 0.6733 \\
Qwen-VL-7B & 0.6390 & 0.8272 & 0.4508 & 0.8076 & 0.8413 & 0.8223 & 0.9136 & 0.4009 & 0.1355 & 0.9430 & 0.3100 & 0.8183 & 0.3970 \\
Qwen-VL-32B & 0.6458 & 0.6762 & 0.5689 & 0.6862 & 0.6941 & 0.6209 & 0.5776 & 0.4948 & 0.1862 & 0.8023 & 0.7618 & 0.7708 & 0.9001 \\
Qwen-VL-72B & 0.6724 & 0.6752 & 0.6925 & 0.7131 & 0.6731 & 0.6921 & 0.7804 & 0.2620 & 0.2447 & 0.8739 & 0.7952 & 0.7729 & 0.8734 \\
Intern-VL3-8B & 0.6331 & 0.4617 & 0.6452 & 0.7577 & 0.7751 & 0.3487 & 0.3329 & 0.3082 & 0.5180 & 0.9486 & 0.6951 & 0.8571 & 0.9491 \\
Intern-VL3-14B & 0.6883 & 0.7406 & 0.3826 & 0.7563 & 0.7601 & 0.7706 & 0.8594 & 0.5276 & 0.1912 & 0.8309 & 0.8304 & 0.8373 & 0.7729 \\
GPT-4o & 0.7031 & 0.7588 & 0.6789 & 0.6597 & 0.4688 & 0.8129 & 0.8190 & 0.8237 & 0.3292 & 0.7506 & 0.7344 & 0.7857 & 0.8160 \\
GPT-5 & \textbf{0.7252} & 0.8378 & 0.7832 & 0.7275 & 0.6510 & 0.6953 & 0.8375 & 0.4000 & 0.5237 & 0.7472 & 0.7419 & 0.9023 & 0.8551 \\
\midrule
\rowcolor{black!5}
\multicolumn{14}{c}{\textbf{Reasoning-augmented Models}} \\
\midrule
GPT-o1 & 0.6705 & 0.7384 & 0.7531 & 0.5522 & 0.3398 & 0.7149 & 0.5439 & 0.7427 & 0.6750 & 0.7049 & 0.7266 & 0.7518 & 0.8030 \\
GPT-o3 & \underline{0.7105} & 0.7450 & 0.7597 & 0.6588 & 0.3964 & 0.7715 & 0.7005 & 0.7993 & 0.6316 & 0.7615 & 0.7332 & 0.8084 & 0.7596 \\
Gemini-2.5-Pro & 0.6368 & 0.7355 & 0.6924 & 0.5936 & 0.5089 & 0.7333 & 0.6495 & 0.6776 & 0.5604 & 0.5888 & 0.6997 & 0.5217 & 0.6803 \\
\midrule
\rowcolor{black!5}
\multicolumn{14}{c}{\textbf{Expert Models for Image Aesthetics Assessment}} \\
\midrule
AesExpert-7B & 0.4056 & 0.5025 & 0.4142 & 0.3317 & 0.4636 & 0.3017 & 0.4147 & 0.3253 & 0.2318 & 0.2670 & 0.5073 & 0.6166 & 0.4904 \\
UNIAA-LLaVA & 0.2900 & 0.2393 & 0.2120 & 0.2471 & 0.2207 & 0.2733 & 0.3316 & 0.3041 & 0.3073 & 0.2893 & 0.5266 & 0.2410 & 0.2879 \\
\bottomrule
\end{tabular}%
}
\label{tab: Aesthetic Judgment}
\end{table*}

\begin{table*}[t]
\centering
\caption{Evaluation on region selection task. Overall acc is the average value of all indicators. The best and second-best results are highlighted in \textbf{bold} and \underline{underlined}, respectively.} 
\label{tab:model_design_evaluation_new}
\resizebox{\textwidth}{!}{%
\begin{tabular}{c c cccc cccc cc cc}
\toprule
\multirow{2}{*}{\textbf{Model}} & \multirow{2}{*}{\textbf{Overall Acc}} & \multicolumn{4}{c}{\textbf{Layout}} & \multicolumn{4}{c}{\textbf{Color}} & \multicolumn{2}{c}{\textbf{Font}} & \multicolumn{2}{c}{\textbf{Graphics}} \\
\cmidrule(lr){3-6} \cmidrule(lr){7-10} \cmidrule(lr){11-12} \cmidrule(lr){13-14}
& & balance & layering & whitespace & alignment & harmony & contrast & appeal & psycholoy & legibility & hierarchy & quality & relevance \\
\midrule
\rowcolor{black!5}
\multicolumn{14}{c}{\textbf{Non-reasoning Models}} \\
\midrule
LLaVA-13B & 0.6065 & 0.5823 & 0.5713 & 0.5612 & 0.5918 & 0.6171 & 0.5166 & 0.6319 & 0.6856 & 0.6130 & 0.7003 & 0.6329 & 0.5745 \\
 Qwen-VL-7B (Base) & 0.5795 & 0.5128 & 0.5370 & 0.5748 & 0.5443 & 0.5822 & 0.5433 & 0.5412 & 0.6384 & 0.5974 & 0.6379 & 0.6258 & 0.6190 \\
Qwen-VL-32B & 0.6311 & 0.5933 & 0.5397 & 0.5012 & 0.5252 & 0.5678 & 0.7065 & 0.6367 & 0.5735 & 0.7833 & 0.6278 & 0.7650 & 0.7533 \\
Qwen-VL-72B & 0.6626 & 0.5105 & 0.5839 & 0.4547 & 0.5348 & 0.5977 & 0.7225 & 0.6495 & 0.7940 & 0.7728 & 0.7934 & 0.7360 & 0.8015 \\
Intern-VL3-8B & 0.5799 & 0.5242 & 0.4948 & 0.5606 & 0.5363 & 0.5527 & 0.5568 & 0.5805 & 0.6342 & 0.6583 & 0.6157 & 0.6031 & 0.6415 \\
Intern-VL3-14B & 0.6378 & 0.5872 & 0.5204 & 0.5945 & 0.5745 & 0.6282 & 0.6997 & 0.6419 & 0.7034 & 0.7244 & 0.6870 & 0.6109 & 0.6818 \\
GPT-4o & \underline{0.6745} & 0.4714 & 0.4894 & 0.5007 & 0.6011 & 0.7406 & 0.8591 & 0.7166 & 0.8135 & 0.6633 & 0.9080 & 0.6444 & 0.6865 \\
GPT-5 & \textbf{0.6989} & 0.6484 & 0.5929 & 0.6630 & 0.6396 & 0.7229 & 0.7214 & 0.7510 & 0.6847 & 0.8038 & 0.7565 & 0.6953 & 0.7070 \\
\midrule
\rowcolor{black!5}
\multicolumn{14}{c}{\textbf{Reasoning-augmented Models}} \\
\midrule
GPT-o1 & 0.6347 & 0.6319 & 0.5746 & 0.6397 & 0.5934 & 0.6178 & 0.6323 & 0.6880 & 0.7734 & 0.6320 & 0.7092 & 0.5936 & 0.5305 \\
GPT-o3 & 0.6581 & 0.6483 & 0.6263 & 0.5325 & 0.3653 & 0.8272 & 0.5486 & 0.7113 & 0.7981 & 0.6586 & 0.7744 & 0.6601 & 0.7466 \\
Gemini-2.5-Pro & 0.6100 & 0.6810 & 0.6981 & 0.6096 & 0.2992 & 0.6050 & 0.6538 & 0.6678 & 0.6050 & 0.6164 & 0.6539 & 0.5696 & 0.6605 \\
\midrule
\rowcolor{black!5}
\multicolumn{14}{c}{\textbf{Expert Models for Image Aesthetics Assessment}} \\
\midrule
AesExpert-7b & 0.2883 & 0.2954 & 0.2280 & 0.3174 & 0.2631 & 0.3426 & 0.2646 & 0.3176 & 0.3176 & 0.3176 & 0.2588 & 0.2765 & 0.2602 \\
UNIAA-LLaVA & 0.2418 & 0.1619 & 0.2915 & 0.1552 & 0.4075 & 0.1700 & 0.1516 & 0.1760 & 0.2479 & 0.2777 & 0.3796 & 0.2861 & 0.1968 \\
\bottomrule
\end{tabular}%
}
\label{tab: Region Selection}
\end{table*}

\section{Experiment}
\subsection{Benchmarking VLMs on AesEval-Bench}
\label{sec: result}
\textbf{Setups.}
We conduct a comprehensive evaluation of 10 VLMs spanning diverse model families and parameter scales. 
For non-reasoning models, we consider open-source representatives such as LLaVA~\citep{liu2023visual}, Qwen2.5-VL~\citep{bai2025qwen2.5}, and Intern-VL3~\citep{zhu2025internvl3}, as well as closed-source GPT models~\citep{jaech2024openai}. 
For reasoning-augmented models, we evaluate GPT-o1, GPT-o3, and Gemini-2.5-Pro~\citep{comanici2025gemini}. 
In addition, we include expert models specifically designed for image aesthetic assessment, namely AesExpert~\citep{huang2024aesbench} and UNIAA-LLAVA~\citep{zhou2024uniaa}. 
All models are evaluated under the same input setting, which consists of a question (see Figure~\ref{fig:benchmark overview}), a design image, and metadata in JSON format.


\textbf{Results.}
The performance of VLMs are evaluated following the protocols introduced in Section~\ref{subsec:eval_protocol}.
Specifically, in addition to reporting scores for each individual indicator, we also provide an overall score computed as the average across all indicators.

\underline{\emph{Aesthetic Judgment.}}
Table~\ref{tab: Aesthetic Judgment} presents the results.
First, among non-reasoning models, GPT-5 achieves the highest performance, with an overall accuracy of 0.7252. This suggests that even the strongest VLMs still struggle with design aesthetics assessment.
Second, reasoning-augmented models do not outperform their non-reasoning counterparts (e.g., GPT-o1~\citep{jaech2024openai} and GPT-o3 vs. GPT-4o~\citep{hurst2024gpt} and GPT-5), indicating that generic reasoning provides little benefit in this domain.
Third, expert models designed for image aesthetics assessment perform worse overall, highlighting a substantial gap between design aesthetics and image aesthetics.
Finally, model performance varies across indicators. For instance, the Qwen-VL series tends to perform better on legibility but worse on psychology compared to other VLMs. 


\underline{\emph{Region Selection.}}
Table~\ref{tab: Region Selection} reports the results.
First, VLM performance on this task is generally worse than on aesthetic judgment, likely because it requires not only assessing whether a design is pleasing but also identifying where flaws occur.
Second, consistent with aesthetic judgment, GPT-5 achieves the best performance, while reasoning-augmented models show no clear advantage.
Finally, across model families, larger models (e.g.,  72B) typically outperform smaller ones (e.g., 7B).
We find some model has the phenomenon of overfitting, so we adopt a weighted sum when calculating final score.

\underline{\emph{Precise Localization.}}
As described in Section~\ref{subsec:eval_protocol}, this task consists of two components, each evaluated separately: a choice problem, where the model predicts None if no aesthetic issue exists (Table~\ref{tab: none}), and a bbox prediction problem, where the model outputs the exact bbox of the aesthetic issue (Table~\ref{tab: bbox}).
We exclude some VLMs (e.g., Intern-VL series and small Qwen-VL models) because they failed to produce meaningful bbox predictions.
For the choice problem, VLMs achieve reasonable performance, with the best model reaching an overall score of 0.6090.
For the bbox prediction problem, even the best-performing model, GPT-5, scores below 0.20, highlighting the substantial difficulty of precisely localizing aesthetic issues.

\begin{figure}[t]
    \centering
    \begin{subfigure}[b]{0.32\textwidth} 
        \includegraphics[width=\textwidth]{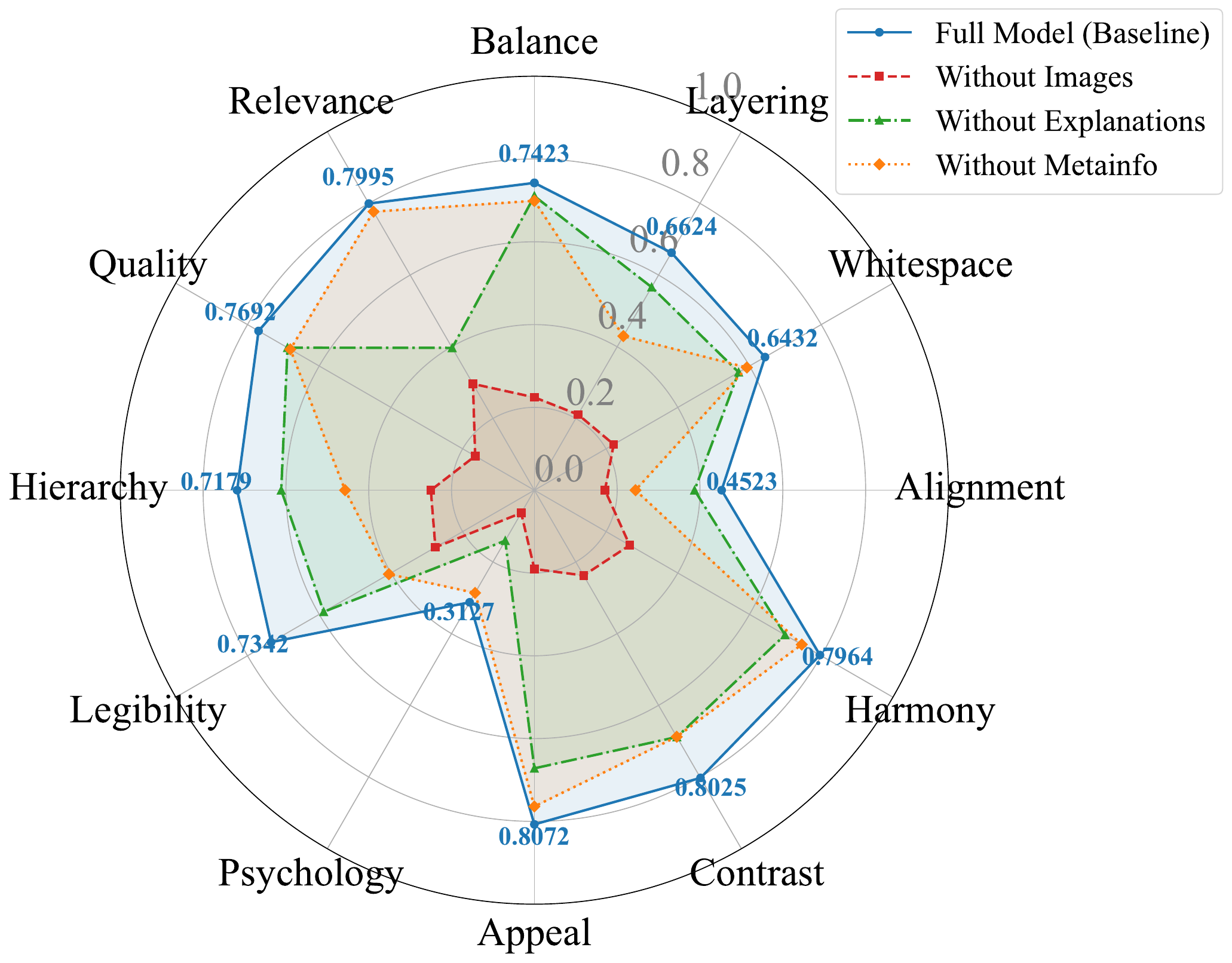}
        \label{fig:ablation-initial}
    \end{subfigure}
    \hfill 
    \begin{subfigure}[b]{0.32\textwidth}
        \includegraphics[width=\textwidth]{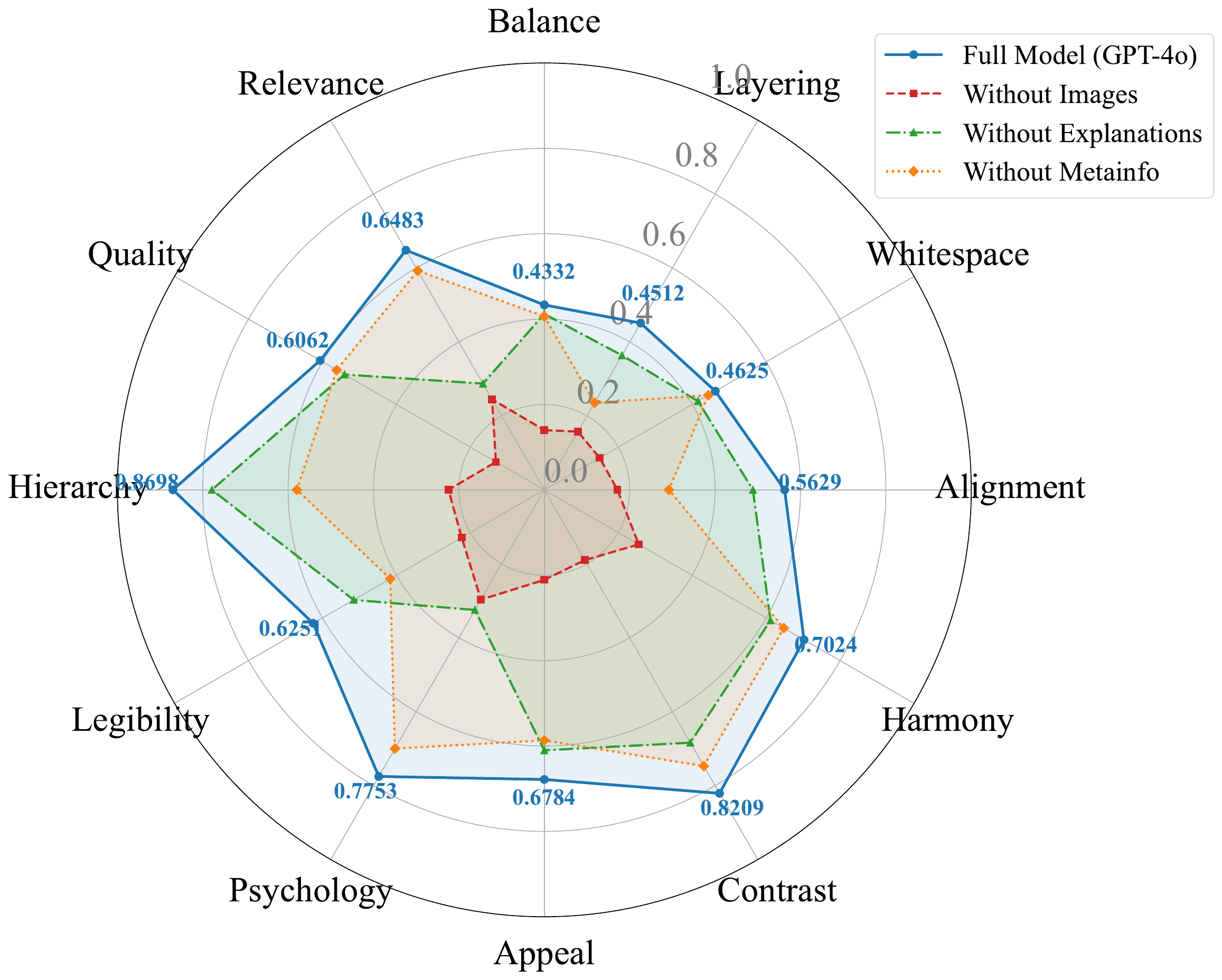}
        \label{fig:ablation-second}
    \end{subfigure}
    \hfill
    \begin{subfigure}[b]{0.32\textwidth}
        \includegraphics[width=\textwidth]{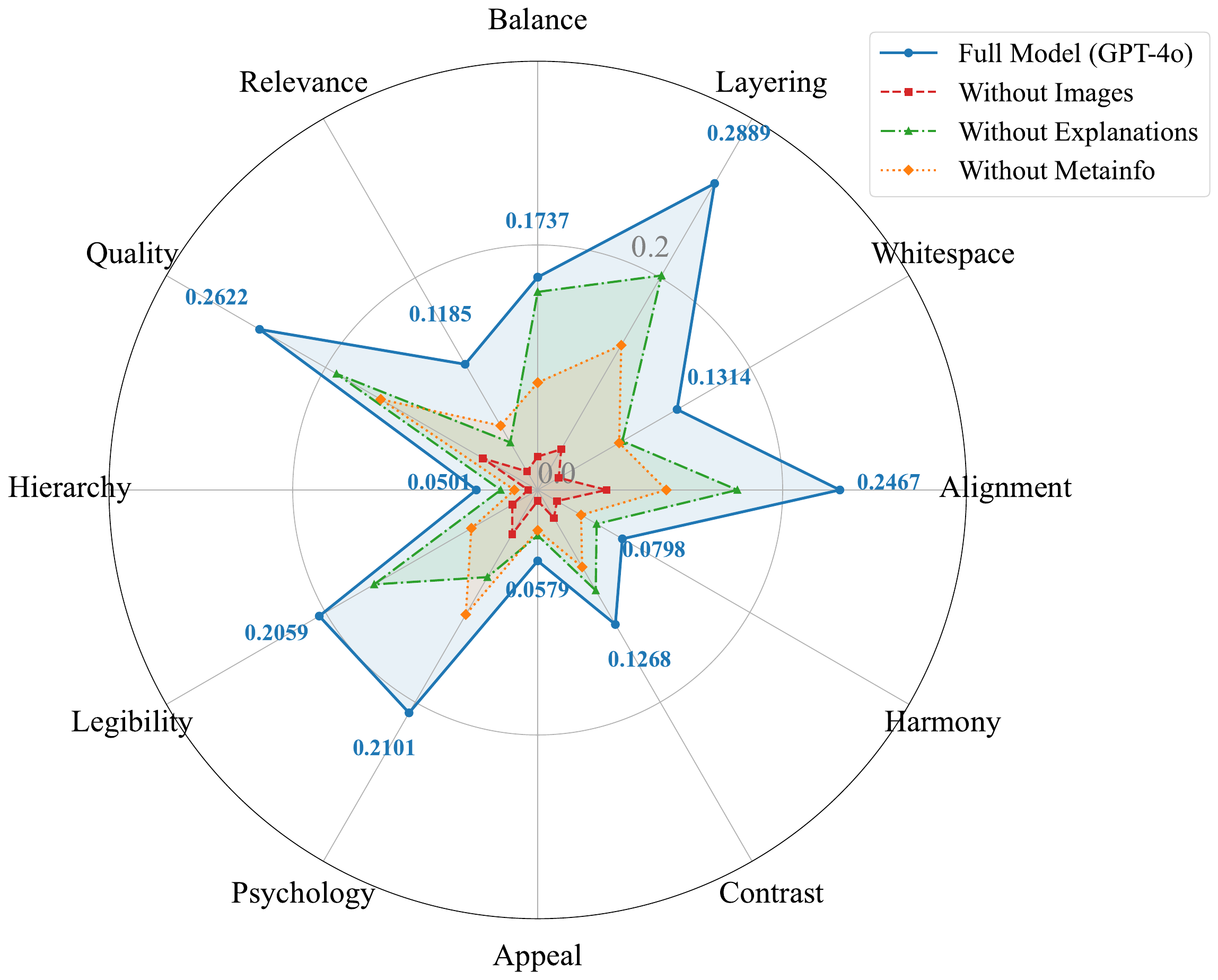}
        \label{fig:ablation-third}
    \end{subfigure}
    \caption{Results for model variants using different input components.}
    \label{fig:ablation}
\end{figure}

\underline{\emph{Discussions on Input Components.}} 
When benchmarking VLMs, the input consists of three components: (1) the question, which includes a detailed explanation of the target indicator; (2) the design image; and (3) metadata in JSON format, containing layout, color, and font information.
We analyze the contribution of each component to model performance using GPT-4o as a representative example.
Figure~\ref{fig:ablation} presents the results, where \emph{Full Model} denotes the setting that uses all three components; \emph{Without Images} removes the design image; \emph{Without Explanation} omits the detailed indicator description; and \emph{Without Metainfo} excludes the metadata.
Our findings reveal three key insights.
First, across all tasks, the design image is indispensable---its removal results in the largest performance drop.
Second, indicator explanations have limited influence for more intuitive indicators (e.g., balance), but they play a crucial role for subjective indicators (e.g., relevance or psychology), where clearer definitions are necessary.
Finally, metadata has the greatest effect on precise localization, where its absence causes a larger decline in performance compared to aesthetic judgment or region selection.
We hypothesize that this is because metadata provides explicit layout information, which aids bbox prediction in localization tasks.


\begin{table*}[t]
\centering
\caption{Evaluation on precise localization task for the choice component where the model should predict None if no aesthetic issues are present. Overall score is the average accuracy of all indicators. The best and second-best results are highlighted in \textbf{bold} and \underline{underlined}, respectively.} 
\label{tab:model_design_evaluation}
\resizebox{\textwidth}{!}{%
\begin{tabular}{cccccccccccccc}
\toprule
\multirow{2}{*}{\textbf{Model}} & \multirow{2}{*}{\textbf{Overall Score}} & \multicolumn{4}{c}{\textbf{Layout}} & \multicolumn{4}{c}{\textbf{Color}} & \multicolumn{2}{c}{\textbf{Font}} & \multicolumn{2}{c}{\textbf{Graphics}} \\
\cmidrule(lr){3-6} \cmidrule(lr){7-10} \cmidrule(lr){11-12} \cmidrule(lr){13-14}
& & balance & layering & whitespace & alignment & harmony & contrast & appeal & psycholoy & legibility & hierarchy & quality & relevance \\
\midrule
\rowcolor{black!5}
\multicolumn{14}{c}{\textbf{Non-reasoning Models}} \\
\midrule
LLaVA-13B & 0.4455 & 0.4523 & 0.6130 & 0.5699 & 0.4714 & 0.3356 & 0.3723 & 0.2898 & 0.2474 & 0.7301 & 0.2453 & 0.5611 & 0.4573 \\
Qwen-VL-7B & 0.5192 & 0.5104 & 0.5839 & 0.4546 & 0.5347 & 0.4376 & 0.5625 & 0.5495 & 0.5339 & 0.4528 & 0.5334 & 0.5360 & 0.5415 \\
GPT-4o & 0.5680 & 0.6417 & 0.2063 & 0.5954 & 0.1679 & 0.8626 & 0.6372 & 0.7200 & 0.6713 & 0.4594 & 0.5164 & 0.7762 & 0.5618 \\
GPT-5 & \textbf{0.6090} & 0.6306 & 0.6142 & 0.6910 & 0.6247 & 0.6057 & 0.6170 & 0.5643 & 0.5989 & 0.5464 & 0.6134 & 0.6217 & 0.5804 \\
\midrule
\rowcolor{black!5}
\multicolumn{14}{c}{\textbf{Reasoning-augmented Models}} \\
\midrule
GPT-o1 & 0.5295 & 0.4628 & 0.4870 & 0.5248 & 0.4943 & 0.5322 & 0.4933 & 0.4912 & 0.5884 & 0.5474 & 0.5879 & 0.6258 & 0.5190 \\
GPT-o3 & 0.5800 & 0.5922 & 0.3868 & 0.5369 & 0.6570 & 0.7554 & 0.6595 & 0.6011 & 0.6985 & 0.7396 & 0.3952 & 0.5311 & 0.4063 \\
Gemini-2.5-Pro & \underline{0.6047} & 0.6319 & 0.5746 & 0.5397 & 0.5934 & 0.6178 & 0.6323 & 0.6880 & 0.7734 & 0.5320 & 0.7092 & 0.4336 & 0.5305 \\
\midrule
\rowcolor{black!5}
\multicolumn{14}{c}{\textbf{Expert Models for Image Aesthetics Assessment}} \\
\midrule
AesExpert-7b & 0.3377 & 0.3229 & 0.3146 & 0.3756 & 0.3172 & 0.4276 & 0.3582 & 0.3146 & 0.3267 & 0.3314 & 0.3803 & 0.3025 & 0.2804 \\
\bottomrule
\end{tabular}%
}
\label{tab: none}
\end{table*}

\begin{table*}[t]   
\centering
\caption{Evaluation on precise localization task for the bbox prediction component where the model should output coordinates of the aesthetic issues. Overall score is the average IoU of all indicators. The best and second-best results are highlighted in \textbf{bold} and \underline{underlined}, respectively.} 
\label{tab:model_design_std}
\resizebox{\textwidth}{!}{%
\begin{tabular}{cccccccccccccc}
\toprule
\multirow{2}{*}{\textbf{Model}} & \multirow{2}{*}{\textbf{Overall Score}} & \multicolumn{4}{c}{\textbf{Layout}} & \multicolumn{4}{c}{\textbf{Color}} & \multicolumn{2}{c}{\textbf{Font}} & \multicolumn{2}{c}{\textbf{Graphics}} \\
\cmidrule(lr){3-6} \cmidrule(lr){7-10} \cmidrule(lr){11-12} \cmidrule(lr){13-14}
& & balance & layering & whitespace & alignment & harmony & contrast & appeal & psycholoy & legibility & hierarchy & quality & relevance \\
\midrule
\rowcolor{black!5}
\multicolumn{14}{c}{\textbf{Non-reasoning Models}} \\
\midrule
LLaVA-13B & 0.0559 & 0.0653 & 0.0080 & 0.0302 & 0.0172 & 0.1024 & 0.0792 & 0.0427 & 0.0303 & 0.0188 & 0.0102 & 0.1510 & 0.1160 \\
 Qwen-VL-7B (Base) &0.0514& 0.0067 & 0.1669 & 0.0101 & 0.0259 & 0.0109 & 0.0036 & 0.0039 & 0.2306 & 0.0012 & 0.0452 & 0.0994 & 0.0063\\ 
GPT-4o & \underline{0.1712} & 0.1822 & 0.2974 & 0.1399 & 0.2552 & 0.0883 & 0.1353 & 0.0664 & 0.2186 & 0.2144 & 0.0586 & 0.2707 & 0.1270 \\
GPT-5 & \textbf{0.1993} & 0.1866 & 0.1546 & 0.2077 & 0.1613 & 0.1829 & 0.1525 & 0.1529 & 0.3348 & 0.1745 & 0.1835 & 0.2667 & 0.2338 \\
\midrule
\rowcolor{black!5}
\multicolumn{14}{c}{\textbf{Reasoning-augmented Models}} \\
\midrule
O1 & 0.1286 & 0.0907 & 0.0767 & 0.1412 & 0.1226 & 0.1719 & 0.0204 & 0.1546 & 0.1236 & 0.1440 & 0.1441 & 0.1912 & 0.1617 \\
O3 & 0.1418 & 0.0619 & 0.1915 & 0.0552 & 0.3075 & 0.0700 & 0.0516 & 0.0760 & 0.1479 & 0.1777 & 0.2796 & 0.1861 & 0.0968 \\
Gemini-2.5-Pro & 0.0977 & 0.0518 & 0.1620 & 0.1052 & 0.0710 & 0.0760 & 0.0487 & 0.0490 & 0.2257 & 0.0963 & 0.0903 & 0.0945 & 0.1014 \\
\midrule
\rowcolor{black!5}
\multicolumn{14}{c}{\textbf{Expert Models for Image Aesthetics Assessment}} \\
\midrule
AesExpert-7b & 0.0327 & 0.0440 & 0.0203 & 0.0093 & 0.1084 & 0.0861 & 0.0063 & 0.0049 & 0.0118 & 0.0001 & 0.0348 & 0.0601 & 0.0067 \\
\bottomrule
\end{tabular}%
}
\label{tab: bbox}
\end{table*}


\subsection{Fine-tuning VLMs with AesEval-Train}

\textbf{Setups.}
We construct the training set following the pipeline described in Section~\ref{sec:train_data}, resulting in 30k question–answer pairs.
In our experiments, we use Qwen2.5-VL-7B-Instruct~\citep{bai2025qwen2.5} as a representative model and adopt full-parameter finetuning on the constructed dataset.
The learning rate is set to 1e-6, with a cosine scheduler and a 3$\%$ warmup ratio.
For computational efficiency, training is performed with bfloat16 mixed precision and FlashAttention-2~\citep{dao2022flashattention}.
The vision encoder is kept frozen, while the language model parameters are tuned. 


\textbf{Main Results.}
Table~\ref{tab: ablation} presents the results, where \emph{Qwen-VL-7B (Base)} denotes the base model without finetuning, and \emph{+AesEval-Train} refers to the model finetuned on our constructed training set.
First, across all three tasks, finetuning with AesEval-Train yields substantial performance improvements.
Moreover, on aesthetic judgment, the finetuned model surpasses even the largest Qwen-VL variant (72B parameters), and on precise localization, it outperforms GPT-5 despite the latter having far more parameters.
These results demonstrate that our proposed pipeline effectively constructs training data that significantly enhances model performance.

\textbf{Ablation Studies.}
We investigate the impact of different data recipes on model performance.
Table~\ref{tab: ablation} reports the results, where \emph{-Reasoning Path} denotes training with plain question–answer pairs without the proposed indicator-grounded reasoning, and \emph{-Positive Samples} denotes training only on flawed designs.
We observe that \emph{-Reasoning Path} still improves performance across all three tasks, suggesting that incorporating domain-specific knowledge of design aesthetics is beneficial.
However, its performance remains notably lower than that of the full variant with reasoning paths (\emph{+AesEval-Train}), underscoring the effectiveness of indicator-grounded reasoning.
In addition, \emph{-Positive Samples} performs worse than both \emph{+AesEval-Train} and \emph{-Reasoning Path}, highlighting the importance of maintaining label balance in the training set.


\begin{table*}[t] 
 \centering 
 \caption{Results and ablation study of fine-tuning VLMs using our constructed training set.}
 \resizebox{\textwidth}{!}{%
 \begin{tabular}{cccccccccccccccc} 
 \toprule 
 \multirow{2}{*}{\textbf{Model Setting}} & \multirow{2}{*}{\textbf{Overall Score}} & \multirow{2}{*}{\textcolor{black}{\textbf{Overall Gain}}}& \multirow{2}{*}{\textcolor{black}{\textbf{Overall Rank}}} & \multicolumn{4}{c}{\textbf{Layout}} & \multicolumn{4}{c}{\textbf{Color}} & \multicolumn{2}{c}{\textbf{Font}} & \multicolumn{2}{c}{\textbf{Graphics}} \\ 
 \cmidrule(lr){5-8} \cmidrule(lr){9-12} \cmidrule(lr){13-14} \cmidrule(lr){15-16} 
 & & & & balance & layering & whitespace & alignment & harmony & contrast & appeal & psycholoy & legibility & hierarchy & quality & relevance \\ 
 \midrule 
 \rowcolor{black!5} 
 \multicolumn{16}{c}{\textbf{Aesthetic Judgment (Accuracy)}} \\ 
 \midrule 
 Qwen-VL-7B (Base) &  0.6390  & \textcolor{black}{-} & \textcolor{black}{9} & 0.8272 & 0.4508 & 0.8076 & 0.8413 & 0.8223 & 0.9136 & 0.4009 & 0.1355 & 0.9430 & 0.3100 & 0.8183 & 0.3970 \\
  + AesEval-Train & 0.6987 &\textcolor{black}{+ 5.97\%} & \textcolor{black}{4}& 0.7123 & 0.6789 & 0.7215 & 0.6868 & 0.7031 & 0.6654 & 0.7329 & 0.6577 & 0.7436 & 0.6482 & 0.7096 & 0.7244 \\
  - Reasoning Path & 0.6576&\textcolor{black}{-} & \textcolor{black}{-}& 0.6511 & 0.6589 & 0.6413 & 0.6687 & 0.6309 & 0.6791 & 0.6207 & 0.6893 & 0.6115 & 0.6985 & 0.6508 & 0.6904\\ 
  - Positive Samples &0.2072&\textcolor{black}{-} &\textcolor{black}{-} & 0.2101 & 0.1999 & 0.2202 & 0.1893 & 0.2058 & 0.2147 & 0.1955 & 0.2246 & 0.1855 & 0.2296 & 0.2004 & 0.2108  \\ 
 \midrule 
 \rowcolor{black!5} 
 \multicolumn{16}{c}{\textbf{Region Selection (Accuracy)}} \\ 
 \midrule 
 Qwen-VL-7B (Base) & 0.5795 &\textcolor{black}{-} &\textcolor{black}{10}& 0.5128 & 0.5370 & 0.5748 & 0.5443 & 0.5822 & 0.5433 & 0.5412 & 0.6384 & 0.5974 & 0.6379 & 0.6258 & 0.6190 \\
  + AesEval-Train &0.6065 &\textcolor{black}{+ 2.70\%}& \textcolor{black}{8}&  0.5827 & 0.6108 & 0.5963 & 0.6289 & 0.5714 & 0.6236 & 0.6541 & 0.6412 & 0.6389 & 0.6487 & 0.5899 & 0.5915 \\
  - Reasoning Path & 0.5795 &\textcolor{black}{-} &\textcolor{black}{-}& 0.5732 & 0.5279 & 0.5697 & 0.5322 & 0.5741 & 0.5322 & 0.5341 & 0.6283 & 0.5953 & 0.6318 & 0.6667 & 0.5885 \\ 
  - Positive Samples &0.5327&\textcolor{black}{-} &\textcolor{black}{-} & 0.5089 & 0.5411 & 0.5257 & 0.5543 & 0.5013 & 0.5587 & 0.5291 & 0.5709 & 0.5212 & 0.5788 & 0.5146 & 0.4878 \\ 
 \midrule 
 \rowcolor{black!5} 
 \multicolumn{16}{c}{\textbf{Precise Localization (IoU)}} \\ 
 \midrule 
 Qwen-VL-7B (Base) &0.0514&\textcolor{black}{-} &\textcolor{black}{8} & 0.0067 & 0.1669 & 0.0101 & 0.0259 & 0.0109 & 0.0036 & 0.0039 & 0.2306 & 0.0012 & 0.0452 & 0.0994 & 0.0063\\ 
  + AesEval-Train & 0.2231 &\textcolor{black}{+ 17.17\%} &\textcolor{black}{1}& 0.2518 & 0.1982 & 0.3103 & 0.0857 & 0.2204 & 0.2846 & 0.1552 & 0.3901 & 0.0607 & 0.2313 & 0.1152 & 0.2745\\ 
  - Reasoning Path &0.0782 &\textcolor{black}{-} &\textcolor{black}{-}& 0.1523 & 0.0211 & 0.1987 & 0.0095 & 0.0750 & 0.1204 & 0.0348 & 0.0812 & 0.0159 & 0.0555 & 0.0601 & 0.1139\\ 
  - Positive Samples &0.0641 &\textcolor{black}{-} &\textcolor{black}{-}& 0.1866 & 0.1546 & 0.2077 & 0.1613 & 0.1829 & 0.1525 & 0.1529 & 0.3348 & 0.1745 & 0.1835 & 0.2667 & 0.2338 \\
 \bottomrule 
 \end{tabular}%
 } 
 \label{tab: ablation}
 \end{table*}

\section{Conclusion}
In this work, we introduce AesEval-Bench for design aesthetics assessment, which spans four dimensions, twelve indicators and three quantifiable tasks.
Based on it, we systematically evaluate proprietary, open-source and reasoning-augmented VLMs, revealing clear gaps in design aesthetics assessment.
Furthermore, we construct a training dataset to fine-tune VLMs for this domain.
Experiments show that this dataset significantly improves model performance across all tasks.

\textbf{Limitations.}
First, as Crello serves as the source dataset, the benchmark does not cover all types of graphic design, such as infographics or mobile UIs. 
\textcolor{black}{Second, a fully disentangled taxonomy for indicators is not yet available. Should a more rigorously disentangled taxonomy be proposed in the future, it would be valuable to adopt it and update our indicators accordingly.}
Third, highly subjective aspects of design, such as creativity, are not included. 
Finally, leveraging reinforcement learning to further enhance reasoning capabilities is left for future exploration.

\bibliography{iclr2026_conference}

@article{zhang2025latent,
  title={Latent Sketchpad: Sketching Visual Thoughts to Elicit Multimodal Reasoning in MLLMs},
  author={Zhang, Huanyu and Wu, Wenshan and Li, Chengzu and Shang, Ning and Xia, Yan and Huang, Yangyu and Zhang, Yifan and Dong, Li and Zhang, Zhang and Wang, Liang and others},
  journal={arXiv preprint arXiv:2510.24514},
  year={2025}
}

@article{lin2025perceive,
  title={Perceive Anything: Recognize, Explain, Caption, and Segment Anything in Images and Videos},
  author={Lin, Weifeng and Wei, Xinyu and An, Ruichuan and Ren, Tianhe and Chen, Tingwei and Zhang, Renrui and Guo, Ziyu and Zhang, Wentao and Zhang, Lei and Li, Hongsheng},
  journal={arXiv preprint arXiv:2506.05302},
  year={2025}
}

@article{yang2025qwen3,
  title={Qwen3 technical report},
  author={Yang, An and Li, Anfeng and Yang, Baosong and Zhang, Beichen and Hui, Binyuan and Zheng, Bo and Yu, Bowen and Gao, Chang and Huang, Chengen and Lv, Chenxu and others},
  journal={arXiv preprint arXiv:2505.09388},
  year={2025}
}

@inproceedings{wangwiwattana2024design,
  title={Design Through AI Eyes: Automating Aesthetic Assessments for Learning Graphic Design Concepts},
  author={Wangwiwattana, Chatchai and Meeyen, Areerat},
  booktitle={International Conference on Multi-disciplinary Trends in Artificial Intelligence},
  pages={336--347},
  year={2024},
  organization={Springer}
}

@inproceedings{yamaguchi2021canvasvae,
  title={Canvasvae: Learning to generate vector graphic documents},
  author={Yamaguchi, Kota},
  booktitle={Proceedings of the IEEE/CVF International Conference on Computer Vision},
  pages={5481--5489},
  year={2021}
}

@article{sarch2025grounded,
  title={Grounded Reinforcement Learning for Visual Reasoning},
  author={Sarch, Gabriel and Saha, Snigdha and Khandelwal, Naitik and Jain, Ayush and Tarr, Michael J and Kumar, Aviral and Fragkiadaki, Katerina},
  journal={arXiv preprint arXiv:2505.23678},
  year={2025}
}

@inproceedings{lou2022designeva,
  title={DesignEva: A Design-Supported Tool with Multi-faceted Perceptual Evaluation},
  author={Lou, Yun and Gao, Weiyue and Chen, Pei and Liu, Xuanhui and Yang, Changyuan and Sun, Lingyun},
  booktitle={International Conference on Human-Computer Interaction},
  pages={508--519},
  year={2022},
  organization={Springer}
}

@article{dao2022flashattention,
  title={Flashattention: Fast and memory-efficient exact attention with io-awareness},
  author={Dao, Tri and Fu, Dan and Ermon, Stefano and Rudra, Atri and R{\'e}, Christopher},
  journal={Advances in neural information processing systems},
  volume={35},
  pages={16344--16359},
  year={2022}
}

@article{bai2025qwen2.5,
  title={Qwen2. 5-vl technical report},
  author={Bai, Shuai and Chen, Keqin and Liu, Xuejing and Wang, Jialin and Ge, Wenbin and Song, Sibo and Dang, Kai and Wang, Peng and Wang, Shijie and Tang, Jun and others},
  journal={arXiv preprint arXiv:2502.13923},
  year={2025}
}

@article{lu2020computational,
  title={Computational aesthetics of fine art paintings: The state of the art and outlook},
  author={Lu, Yue and Guo, Chao and Lin, Yilun and Zhuo, Fan and Wang, FY},
  journal={Acta Automatica Sinica},
  volume={46},
  number={11},
  pages={2239--2259},
  year={2020}
}

@article{shao2024visual,
  title={Visual cot: Advancing multi-modal language models with a comprehensive dataset and benchmark for chain-of-thought reasoning},
  author={Shao, Hao and Qian, Shengju and Xiao, Han and Song, Guanglu and Zong, Zhuofan and Wang, Letian and Liu, Yu and Li, Hongsheng},
  journal={Advances in Neural Information Processing Systems},
  volume={37},
  pages={8612--8642},
  year={2024}
}

@inproceedings{mccormack2020understanding,
  title={Understanding aesthetic evaluation using deep learning},
  author={McCormack, Jon and Lomas, Andy},
  booktitle={International conference on computational intelligence in music, sound, art and design (part of EvoStar)},
  pages={118--133},
  year={2020},
  organization={Springer}
}

@article{sun2024visual,
  title={Visual agents as fast and slow thinkers},
  author={Sun, Guangyan and Jin, Mingyu and Wang, Zhenting and Wang, Cheng-Long and Ma, Siqi and Wang, Qifan and Geng, Tong and Wu, Ying Nian and Zhang, Yongfeng and Liu, Dongfang},
  journal={arXiv preprint arXiv:2408.08862},
  year={2024}
}

@article{wu2025grounded,
  title={Grounded chain-of-thought for multimodal large language models},
  author={Wu, Qiong and Yang, Xiangcong and Zhou, Yiyi and Fang, Chenxin and Song, Baiyang and Sun, Xiaoshuai and Ji, Rongrong},
  journal={arXiv preprint arXiv:2503.12799},
  year={2025}
}

@article{cao2025ground,
  title={Ground-R1: Incentivizing Grounded Visual Reasoning via Reinforcement Learning},
  author={Cao, Meng and Zhao, Haoze and Zhang, Can and Chang, Xiaojun and Reid, Ian and Liang, Xiaodan},
  journal={arXiv preprint arXiv:2505.20272},
  year={2025}
}

@misc{luo2025drvhierarchicalperceptiontemporalcognitionframework,
      title={Dr.V: A Hierarchical Perception-Temporal-Cognition Framework to Diagnose Video Hallucination by Fine-grained Spatial-Temporal Grounding}, 
      author={Meng Luo and Shengqiong Wu and Liqiang Jing and Tianjie Ju and Li Zheng and Jinxiang Lai and Tianlong Wu and Xinya Du and Jian Li and Siyuan Yan and Jiebo Luo and William Yang Wang and Hao Fei and Mong-Li Lee and Wynne Hsu},
      year={2025},
      eprint={2509.11866},
      archivePrefix={arXiv},
      primaryClass={cs.CV},
      url={https://arxiv.org/abs/2509.11866}, 
}

@inproceedings{zhangmme,
  title={MME-RealWorld: Could Your Multimodal LLM Challenge High-Resolution Real-World Scenarios that are Difficult for Humans?},
  author={Zhang, YiFan and Zhang, Huanyu and Tian, Haochen and Fu, Chaoyou and Zhang, Shuangqing and Wu, Junfei and Li, Feng and Wang, Kun and Wen, Qingsong and Zhang, Zhang and others},
  booktitle={The Thirteenth International Conference on Learning Representations},
  year={2025}
}

@article{li2024survey,
  title={A survey on benchmarks of multimodal large language models},
  author={Li, Jian and Lu, Weiheng and Fei, Hao and Luo, Meng and Dai, Ming and Xia, Min and Jin, Yizhang and Gan, Zhenye and Qi, Ding and Fu, Chaoyou and others},
  journal={arXiv preprint arXiv:2408.08632},
  year={2024}
}

@article{li2025imagine,
  title={Imagine while reasoning in space: Multimodal visualization-of-thought},
  author={Li, Chengzu and Wu, Wenshan and Zhang, Huanyu and Xia, Yan and Mao, Shaoguang and Dong, Li and Vuli{\'c}, Ivan and Wei, Furu},
  journal={arXiv preprint arXiv:2501.07542},
  year={2025}
}

@article{zhang2025scaling,
  title={Scaling and Beyond: Advancing Spatial Reasoning in MLLMs Requires New Recipes},
  author={Zhang, Huanyu and Li, Chengzu and Wu, Wenshan and Mao, Shaoguang and Zhang, Yifan and Tian, Haochen and Vuli{\'c}, Ivan and Zhang, Zhang and Wang, Liang and Tan, Tieniu and others},
  journal={arXiv preprint arXiv:2504.15037},
  year={2025}
}

@article{deng2017image,
  title={Image aesthetic assessment: An experimental survey},
  author={Deng, Yubin and Loy, Chen Change and Tang, Xiaoou},
  journal={IEEE Signal Processing Magazine},
  volume={34},
  number={4},
  pages={80--106},
  year={2017},
  publisher={IEEE}
}

@article{li2024llava,
  title={Llava-onevision: Easy visual task transfer},
  author={Li, Bo and Zhang, Yuanhan and Guo, Dong and Zhang, Renrui and Li, Feng and Zhang, Hao and Zhang, Kaichen and Zhang, Peiyuan and Li, Yanwei and Liu, Ziwei and others},
  journal={arXiv preprint arXiv:2408.03326},
  year={2024}
}

@article{hurst2024gpt,
  title={Gpt-4o system card},
  author={Hurst, Aaron and Lerer, Adam and Goucher, Adam P and Perelman, Adam and Ramesh, Aditya and Clark, Aidan and Ostrow, AJ and Welihinda, Akila and Hayes, Alan and Radford, Alec and others},
  journal={arXiv preprint arXiv:2410.21276},
  year={2024}
}

@article{lin2024draw,
  title={Draw-and-understand: Leveraging visual prompts to enable mllms to comprehend what you want},
  author={Lin, Weifeng and Wei, Xinyu and An, Ruichuan and Gao, Peng and Zou, Bocheng and Luo, Yulin and Huang, Siyuan and Zhang, Shanghang and Li, Hongsheng},
  journal={arXiv preprint arXiv:2403.20271},
  year={2024}
}

@article{an2024mc,
  title={Mc-llava: Multi-concept personalized vision-language model},
  author={An, Ruichuan and Yang, Sihan and Lu, Ming and Zhang, Renrui and Zeng, Kai and Luo, Yulin and Cao, Jiajun and Liang, Hao and Chen, Ying and She, Qi and others},
  journal={arXiv preprint arXiv:2411.11706},
  year={2024}
}

@article{huang2024aesbench,
  title={Aesbench: An expert benchmark for multimodal large language models on image aesthetics perception},
  author={Huang, Yipo and Yuan, Quan and Sheng, Xiangfei and Yang, Zhichao and Wu, Haoning and Chen, Pengfei and Yang, Yuzhe and Li, Leida and Lin, Weisi},
  journal={arXiv preprint arXiv:2401.08276},
  year={2024}
}

@article{li2009aesthetic,
  title={Aesthetic visual quality assessment of paintings},
  author={Li, Congcong and Chen, Tsuhan},
  journal={IEEE Journal of selected topics in Signal Processing},
  volume={3},
  number={2},
  pages={236--252},
  year={2009},
  publisher={IEEE}
}

@article{jiang2025multimodal,
  title={Multimodal LLMs Can Reason about Aesthetics in Zero-Shot},
  author={Jiang, Ruixiang and Chen, Changwen},
  journal={arXiv preprint arXiv:2501.09012},
  year={2025}
}

@article{wang2024qwen2,
  title={Qwen2-vl: Enhancing vision-language model's perception of the world at any resolution},
  author={Wang, Peng and Bai, Shuai and Tan, Sinan and Wang, Shijie and Fan, Zhihao and Bai, Jinze and Chen, Keqin and Liu, Xuejing and Wang, Jialin and Ge, Wenbin and others},
  journal={arXiv preprint arXiv:2409.12191},
  year={2024}
}

@article{jaech2024openai,
  title={Openai o1 system card},
  author={Jaech, Aaron and Kalai, Adam and Lerer, Adam and Richardson, Adam and El-Kishky, Ahmed and Low, Aiden and Helyar, Alec and Madry, Aleksander and Beutel, Alex and Carney, Alex and others},
  journal={arXiv preprint arXiv:2412.16720},
  year={2024}
}

@article{comanici2025gemini,
  title={Gemini 2.5: Pushing the frontier with advanced reasoning, multimodality, long context, and next generation agentic capabilities},
  author={Comanici, Gheorghe and Bieber, Eric and Schaekermann, Mike and Pasupat, Ice and Sachdeva, Noveen and Dhillon, Inderjit and Blistein, Marcel and Ram, Ori and Zhang, Dan and Rosen, Evan and others},
  journal={arXiv preprint arXiv:2507.06261},
  year={2025}
}

@article{zhu2025internvl3,
  title={Internvl3: Exploring advanced training and test-time recipes for open-source multimodal models},
  author={Zhu, Jinguo and Wang, Weiyun and Chen, Zhe and Liu, Zhaoyang and Ye, Shenglong and Gu, Lixin and Tian, Hao and Duan, Yuchen and Su, Weijie and Shao, Jie and others},
  journal={arXiv preprint arXiv:2504.10479},
  year={2025}
}

@article{liu2023visual,
  title={Visual instruction tuning},
  author={Liu, Haotian and Li, Chunyuan and Wu, Qingyang and Lee, Yong Jae},
  journal={Advances in neural information processing systems},
  volume={36},
  pages={34892--34916},
  year={2023}
}

@article{achiam2023gpt,
  title={Gpt-4 technical report},
  author={Achiam, Josh and Adler, Steven and Agarwal, Sandhini and Ahmad, Lama and Akkaya, Ilge and Aleman, Florencia Leoni and Almeida, Diogo and Altenschmidt, Janko and Altman, Sam and Anadkat, Shyamal and others},
  journal={arXiv preprint arXiv:2303.08774},
  year={2023}
}

@article{lin2023designbench,
  title={Designbench: Exploring and benchmarking dall-e 3 for imagining visual design},
  author={Lin, Kevin and Yang, Zhengyuan and Li, Linjie and Wang, Jianfeng and Wang, Lijuan},
  journal={arXiv preprint arXiv:2310.15144},
  year={2023}
}

@article{lin2024designprobe,
  title={Designprobe: A graphic design benchmark for multimodal large language models},
  author={Lin, Jieru and Huang, Danqing and Zhao, Tiejun and Zhan, Dechen and Lin, Chin-Yew},
  journal={arXiv preprint arXiv:2404.14801},
  year={2024}
}

@inproceedings{haraguchi2024can,
  title={Can GPTs Evaluate Graphic Design Based on Design Principles?},
  author={Haraguchi, Daichi and Inoue, Naoto and Shimoda, Wataru and Mitani, Hayato and Uchida, Seiichi and Yamaguchi, Kota},
  booktitle={SIGGRAPH Asia 2024 Technical Communications},
  pages={1--4},
  year={2024}
}

@article{jung2025ui,
  title={UI-Bench: A Benchmark for Evaluating Design Capabilities of AI Text-to-App Tools},
  author={Jung, Sam and Garcinuno, Agustin and Mateega, Spencer},
  journal={arXiv preprint arXiv:2508.20410},
  year={2025}
}

@article{zhang2024document,
  title={Document parsing unveiled: Techniques, challenges, and prospects for structured information extraction},
  author={Zhang, Qintong and Wang, Bin and Huang, Victor Shea-Jay and Zhang, Junyuan and Wang, Zhengren and Liang, Hao and He, Conghui and Zhang, Wentao},
  journal={arXiv preprint arXiv:2410.21169},
  year={2024}
}

@inproceedings{li2025chemvlm,
  title={Chemvlm: Exploring the power of multimodal large language models in chemistry area},
  author={Li, Junxian and Zhang, Di and Wang, Xunzhi and Hao, Zeying and Lei, Jingdi and Tan, Qian and Zhou, Cai and Liu, Wei and Yang, Yaotian and Xiong, Xinrui and others},
  booktitle={Proceedings of the AAAI Conference on Artificial Intelligence},
  volume={39},
  number={1},
  pages={415--423},
  year={2025}
}

@inproceedings{zhang2025critic,
  title={Critic-v: Vlm critics help catch vlm errors in multimodal reasoning},
  author={Zhang, Di and Lei, Jingdi and Li, Junxian and Wang, Xunzhi and Liu, Yujie and Yang, Zonglin and Li, Jiatong and Wang, Weida and Yang, Suorong and Wu, Jianbo and others},
  booktitle={Proceedings of the Computer Vision and Pattern Recognition Conference},
  pages={9050--9061},
  year={2025}
}

@article{zhou2024uniaa,
  title={Uniaa: A unified multi-modal image aesthetic assessment baseline and benchmark},
  author={Zhou, Zhaokun and Wang, Qiulin and Lin, Bin and Su, Yiwei and Chen, Rui and Tao, Xin and Zheng, Amin and Yuan, Li and Wan, Pengfei and Zhang, Di},
  journal={arXiv preprint arXiv:2404.09619},
  year={2024}
}

@article{luo2025dr,
  title={Dr. V: A Hierarchical Perception-Temporal-Cognition Framework to Diagnose Video Hallucination by Fine-grained Spatial-Temporal Grounding},
  author={Luo, Meng and Wu, Shengqiong and Jing, Liqiang and Ju, Tianjie and Zheng, Li and Lai, Jinxiang and Wu, Tianlong and Du, Xinya and Li, Jian and Yan, Siyuan and others},
  journal={arXiv preprint arXiv:2509.11866},
  year={2025}
}

@article{an2026genius,
  title={GENIUS: Generative Fluid Intelligence Evaluation Suite},
  author={An, Ruichuan and Yang, Sihan and Guo, Ziyu and Dai, Wei and Shen, Zijun and Li, Haodong and Zhang, Renrui and Wei, Xinyu and Li, Guopeng and Wu, Wenshan and others},
  journal={arXiv preprint arXiv:2602.11144},
  year={2026}
}

@article{guo2025video,
  title={Are video models ready as zero-shot reasoners? an empirical study with the mme-cof benchmark},
  author={Guo, Ziyu and Chen, Xinyan and Zhang, Renrui and An, Ruichuan and Qi, Yu and Jiang, Dongzhi and Li, Xiangtai and Zhang, Manyuan and Li, Hongsheng and Heng, Pheng-Ann},
  journal={arXiv preprint arXiv:2510.26802},
  year={2025}
}

@inproceedings{luo2024panosent,
  title={Panosent: A panoptic sextuple extraction benchmark for multimodal conversational aspect-based sentiment analysis},
  author={Luo, Meng and Fei, Hao and Li, Bobo and Wu, Shengqiong and Liu, Qian and Poria, Soujanya and Cambria, Erik and Lee, Mong-Li and Hsu, Wynne},
  booktitle={Proceedings of the 32nd ACM International Conference on Multimedia},
  pages={7667--7676},
  year={2024}
}

@article{zhong2025vcu,
  title={VCU-Bridge: Hierarchical Visual Connotation Understanding via Semantic Bridging},
  author={Zhong, Ming and Wang, Yuanlei and Zhang, Liuzhou and An, Arctanx and Zhang, Renrui and Liang, Hao and Lu, Ming and Shen, Ying and Zhang, Wentao},
  journal={arXiv preprint arXiv:2511.18121},
  year={2025}
}

@inproceedings{singh2025trishul,
  title={Trishul: Towards region identification and screen hierarchy understanding for large vlm based gui agents},
  author={Singh, Kunal and Singh, Shreyas and Khanna, Mukund},
  booktitle={Proceedings of the Computer Vision and Pattern Recognition Conference},
  pages={170--179},
  year={2025}
}

@inproceedings{park2025r,
  title={R-vlm: Region-aware vision language model for precise gui grounding},
  author={Park, Joonhyung and Tang, Peng and Das, Sagnik and Appalaraju, Srikar and Singh, Kunwar Yashraj and Manmatha, R and Ghadar, Shabnam},
  booktitle={Findings of the Association for Computational Linguistics: ACL 2025},
  pages={9669--9685},
  year={2025}
}

@inproceedings{kang2025open,
  title={Open-ended hierarchical streaming video understanding with vision language models},
  author={Kang, Hyolim and Park, Yunsu and Yoo, Youngbeom and Choi, Yeeun and Kim, Seon Joo},
  booktitle={Proceedings of the IEEE/CVF International Conference on Computer Vision},
  pages={20715--20725},
  year={2025}
}
\bibliographystyle{iclr2026_conference}

\newpage
\appendix

\begin{table*}[h!]
\centering
\begin{minipage}{0.9\columnwidth}
\centering
\begin{tcolorbox}[
  colback=VeryLightGray, 
  boxrule=1.2pt
]
\small
\textcolor{black}{\textbf{Dimension-Indicators Prompts:}} \\

\textcolor{purple!50!white}{\textbf{Graphic-Quality:}} Resolution is a measure of image detail. Appropriate resolution ensures the best picture quality and readability. Does element have quality issue?

\textcolor{purple!50!white}{\textbf{Graphic-Relevance:}} Relevance refers to the direct connection between a graphic element and the meaning it conveys. Does element have relevance issue? \\

\textcolor{red!50!white}{\textbf{Color-Harmony:}} Color Harmony refers to the overall coordination, pleasure and beauty of the entire color when there are two or more colors in an image.

\textcolor{red!50!white}{\textbf{Color-Contrast:}} Color contrast refers to the contrasts, oppositions, and differences existing among various colors.

\textcolor{red!50!white}{\textbf{Color-Appreal:}} Color appeal refers to the fact that the selection and combination of colors can attract the attention of the audience.

\textcolor{red!50!white}{\textbf{Color-Psychology:}} Color psychology refers to the idea that color can trigger subjective psychological experiences and influence emotions, feelings, and behaviors.
\\

\textcolor{orange!50!white}{\textbf{Layout-Balance:}} Balance is the distribution of visual weight in design. It can be symmetrical (with equal weights on both sides) or asymmetrical (with unequal weights but still achieving visual balance).

\textcolor{orange!50!white}{\textbf{Layout-Layering:}} Use size, color, contrast, and other visual cues to establish a hierarchical structure of design elements to guide the audience's eyes.

\textcolor{orange!50!white}{\textbf{Layout-Whitespace:}} White space refers to the blank area around the elements in a design. Effective utilization of white space can create balance, visual hierarchy, and clarity.

\textcolor{orange!50!white}{\textbf{Layout-Alignment:}} Alignment refers to the arrangement of design elements relative to each other or a specific axis or grid. Proper alignment creates a sense of order and organization, making the design easier to understand and navigate. 
\\

\textcolor{pink}{\textbf{Font-Hierarchy:}} The presentation of the font has a hierarchical structure, so users can scan the text to obtain key information.

\textcolor{pink}{\textbf{Font-Legibility:}} Legibility refers to the recognition of individual characters and the relationships between them when they are arranged side by side.

\end{tcolorbox} 

\vspace{-2mm}
\caption{Showcase of descriptions of each indicator.}
\label{tab: prompts}
\end{minipage}
\end{table*}

\begin{table*}[h!]
\centering
\begin{minipage}{0.9\columnwidth}
\centering
\begin{tcolorbox}[
  colback=VeryLightGray, 
  boxrule=1.2pt
]
\small
\textcolor{black}{\textbf{Task Reasoning Instructions:}} \\

\textcolor{purple!50!white}{\textbf{Aesthetic Judgment :}} Given the preview image and the element image and element bounding box {bbox}, please reason why the element is not aesthetic in the aspect of {criteria}. Your reason should only contain the analysis about {criteria}.Please think and reasoning. Once the element is mentioned, you should add its bounding box after the element(only bounding box).
\\

\textcolor{red!50!white}{\textbf{Region Selection:}} Given a preview image, an image of an unsightly element and its bounding box {bbox[0]}, and an image of an attractive element and its bounding box {bbox[1]}, analyze why the first element is unsightly in terms of {criteria} based on the relationship between the elements. Your reasoning should only include an analysis of {criteria}. Please think and reason. Once an element is mentioned, you should add its bounding box after the element (bounding box only).
\\

\textcolor{orange!50!white}{\textbf{Precise Localization:}} Given a preview image, an element image, and the element's bounding box {bbox}, analyze why the element is unsightly when viewed from the perspective of {criteria}, based on their relationship. Your reasoning should only include analysis of {criteria}. Think and reason. Once you mention the element, you should follow it with its bounding box (only the bounding box).
\\

\end{tcolorbox} 
\vspace{-2mm}
\caption{Showcase of task reasoning  instructions.}
\label{tab: instructions}
\end{minipage}
\end{table*}

\newpage
\section{Statement of LLM Usage}

Large language models (LLMs) were consulted for technical guidance during implementation and debugging; following the collaborative drafting of the manuscript, we further employed LLMs to refine the prose and enhance the overall exposition.

\section{Prompts and Instructions}
First, Dimension-Indicator Prompts establish a clear set of evaluation criteria. These are organized into four core dimensions: Graphics, Color, Layout, and Font,     each containing specific indicators like Font-Legibility. As shown in Table~\ref{tab: prompts}, every indicator is defined and paired with a guiding question to standardize the analysis.

Second, Task Reasoning Instructions (Table ~\ref{tab: instructions}) provide operational guidance for creating the reasoning paths. They direct the analysis to focus on an element's intrinsic flaws, its relationship with other elements, or its immediate context, while critically mandating the inclusion of the element’s bounding box (bbox) to ground the reasoning in precise spatial evidence.

\section{Additional Related Work}
\textbf{Evaluation for VLMs.}
Recent years have witnessed a surge in benchmarks~\citep{li2024survey} designed to evaluate Vision-Language Models (VLMs), ranging from general-purpose assessments of perception~\citep{luo2024panosent} and reasoning~\citep{guo2025video, zhang2025critic}, which encompass region-level~\citep{lin2024draw, lin2025perceive, park2025r} and hierarchical understanding~\citep{kang2025open,singh2025trishul,zhong2025vcu}, to more specialized evaluations in domains like chemistry~\citep{li2025chemvlm} or image generation~\citep{an2026genius}. 
While existing benchmarks effectively evaluate general capability, they often overlook the nuanced, subjective dimensions of visual understanding. To address this, our work introduces a specialized benchmark focused on the aesthetic dimension, assessing how VLMs interpret artistic quality and visual appeal. 

\textcolor{black}{\section{Data Source Showcase}}

\begin{figure}
    \centering
    \includegraphics[width=1\linewidth]{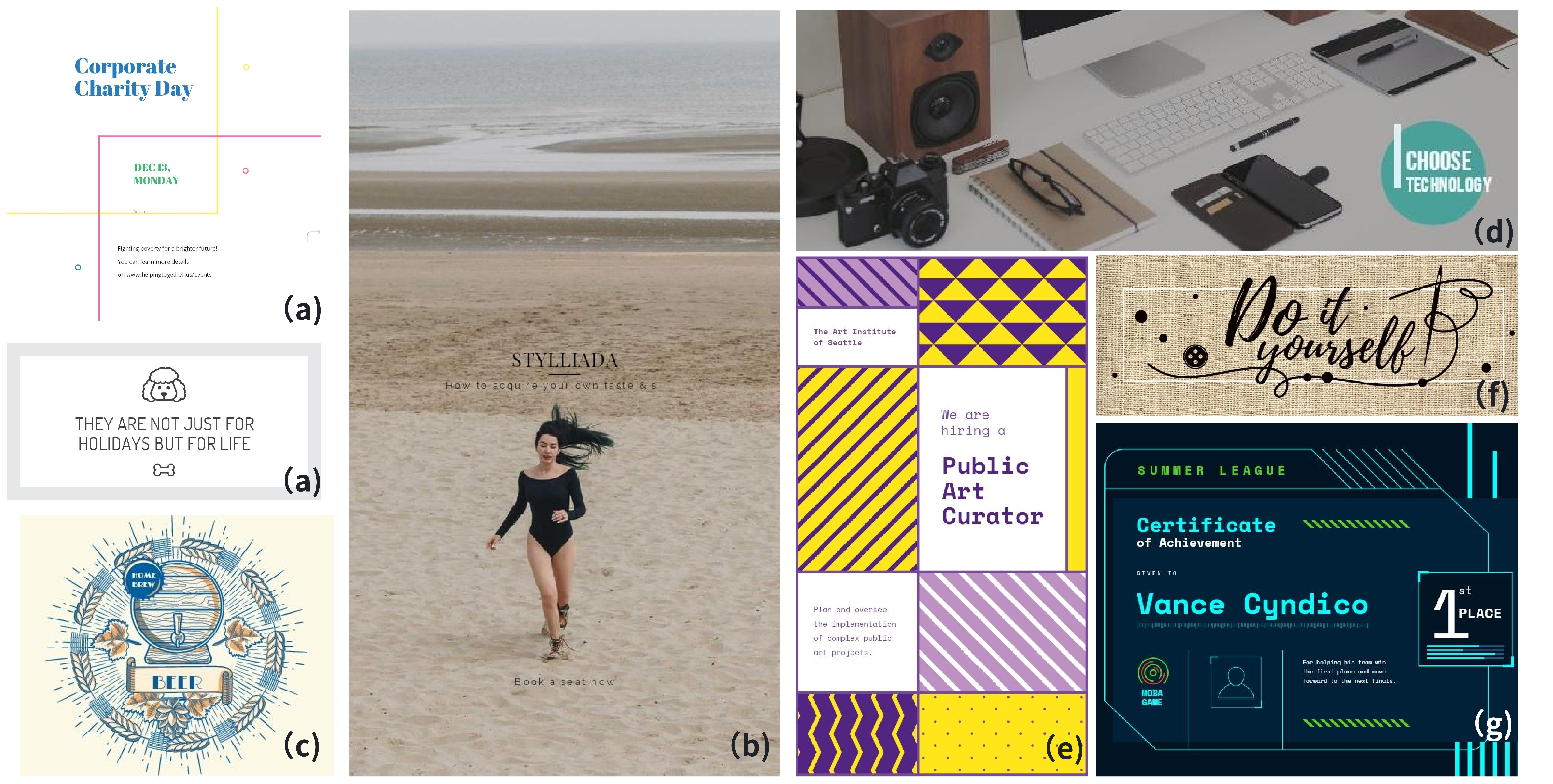}

    \caption{
    \textcolor{black}{\textbf{Diverse design samples sourced from the Crello dataset.} The collection demonstrates a wide spectrum of visual styles and structural layouts, including (a) minimalist typography-centric designs (e.g., "Corporate Charity"), (b) photography-driven fashion editorials featuring real human subjects, (c) vintage illustrations, (d) photorealistic tech mockups, (e) geometric abstract art, (f) textured artistic typography, and (g) cyberpunk-themed certificates. This visual evidence refutes the concern of stylistic homogeneity, confirming the dataset's robust coverage across design domains.
    }}
    \label{fig:crello}
\end{figure}

\textcolor{black}{To address the concern regarding the diversity of the Crello dataset, we provide a visual sampling of its typical data in Fig. ~\ref{fig:crello}. As illustrated, the dataset covers a comprehensive range of design categories and aesthetics, including textual handwritten, photography-driven and diverse art styles designs. This diversity ensures that our benchmark evaluates models on a realistic distribution of graphic design tasks, preventing bias toward any single visual style.}

\textcolor{black}{\section{Flaw Injection Pipeline}}

\begin{algorithm}[h]
\color{black} 
\caption{\color{black}Data Construction Pipeline via Synthetic Perturbation} 
\label{alg:data_construction}
\begin{algorithmic}[1]
\REQUIRE Original Design $D$, Metadata $M$ (in JSON format), Perturbation Library $\mathcal{P}$, Number of perturbed elements $n$.
\ENSURE Perturbed Design $D'$, Updated Metadata $M'$.

\STATE \textbf{Initialization:} $M' \leftarrow M$
\STATE \textbf{Element Selection:} Randomly select a subset of elements $E = \{e_1, e_2, \dots, e_n\}$ from $M'$.

\FORALL{$e_i \in E$}
    \STATE \textbf{Perturbation Selection:} Randomly sample an operation $op \in \mathcal{P}$ applicable to the type of $e_i$.
    \STATE \textbf{Parameter Sampling:} Sample perturbation intensity $\delta$ or target attributes.
    
    \IF{$op$ is \textit{Layout Perturbation}}
        \STATE $e_i.\text{pos} \leftarrow e_i.\text{pos} + \mathcal{U}(-\delta_{pos}, \delta_{pos})$ \COMMENT{Shift position}
        
    \ELSIF{$op$ is \textit{Font Perturbation}}
        \IF{$op$ is Size Change}
            \STATE $e_i.\text{size} \leftarrow e_i.\text{size} + \mathcal{U}(-\delta_{size}, \delta_{size})$
        \ELSIF{$op$ is Font Swap}
            \STATE $e_i.\text{font} \leftarrow \text{Sample}(\text{FontLibrary}) \setminus \{e_i.\text{font}\}$
        \ENDIF
        
    \ELSIF{$op$ is \textit{Color Perturbation}}
        \IF{$op$ is Low Contrast}
            \STATE $e_i.\text{color} \leftarrow \text{SampleNear}(M.\text{background\_color}, \epsilon)$
        \ELSIF{$op$ is High Contrast / Clashing}
            \STATE $e_i.\text{color} \leftarrow \text{Invert}(M.\text{dominant\_color}) + \text{Noise}$
        \ENDIF
        
    \ELSIF{$op$ is \textit{Graphic/Image Perturbation}}
        \IF{$op$ is Replacement}
            \STATE $e_i.\text{src} \leftarrow \text{Sample}(\text{ImageLibrary})$
        \ELSIF{$op$ is Resolution Reduction}
            \STATE $e_i.\text{quality} \leftarrow \text{Downsample}(e_i.\text{src}, \text{factor})$
        \ELSIF{$op$ is Blur}
            \STATE $e_i.\text{effect} \leftarrow \text{GaussianBlur}(e_i.\text{src}, \sigma)$
        \ENDIF
    \ENDIF
    
    \STATE \textbf{Update:} Update element $e_i$ within metadata $M'$.
\ENDFOR

\STATE \textbf{Rendering:} $D' \leftarrow \text{Render}(M')$ \COMMENT{Re-render design using updated JSON}
\RETURN $D', M'$
\end{algorithmic}
\end{algorithm}

\textcolor{black}{\section{Real World Design Data Showcase}}
\begin{figure}[t]
    \centering
    \includegraphics[width=1\linewidth]{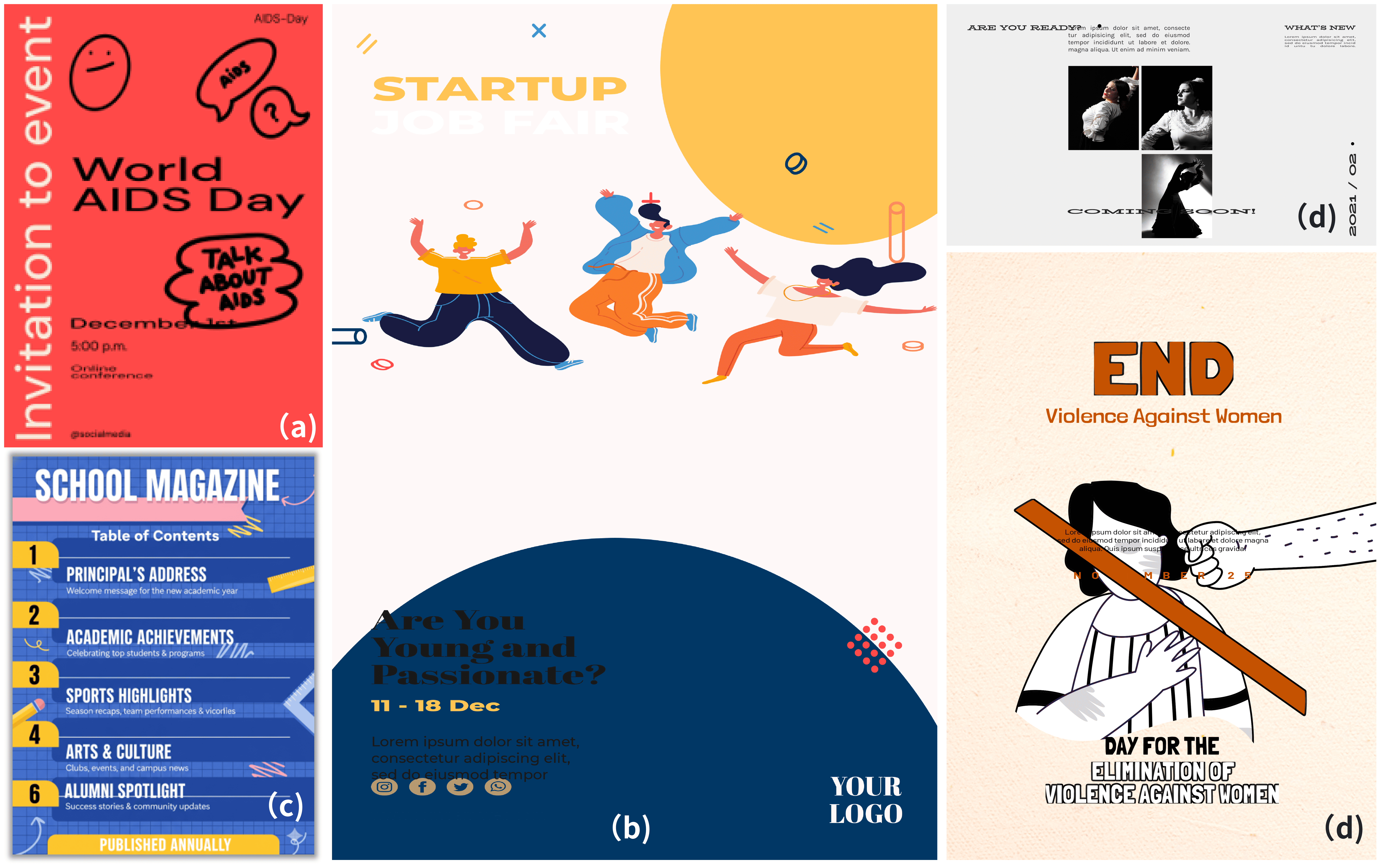}

    \caption{
    \textcolor{black}{Representative samples from real-world flawed cases collected by professional designers. Unlike the synthetically perturbed benchmark, these designs were curated by professional designers to represent authentic aesthetic defects encountered in real-world workflows. The samples exhibit flaws such as (a) visual clutter and inconsistent orientation (e.g., "World AIDS Day"), (b) spatial imbalance and disconnected elements (e.g., "Startup Job Fair"), (c) grid-based alignment errors (e.g., "School Magazine"), and (d) typographic obstruction (e.g., "End Violence"). This dataset serves as a rigorous Out-Of-Distribution benchmark to evaluate model generalization beyond synthetic patterns.
    }}
    \label{fig:real world}
\end{figure}

\textcolor{black}{To validate our model's generalization capabilities beyond the synthetic distributions of Crello, we compiled a distinct Out-Of-Distribution (OOD) test set consisting of real-world flawed designs. As shown in Fig. ~\ref{fig:real world}, these samples were sourced directly from intermediate design drafts and annotated by professional designers, capturing the nuanced and often complex nature of authentic.}

\textcolor{black}{\textbf{Mapping Real-World Flaws to Synthetic Perturbations.} To further validate the design of our data construction pipeline, we analyzed the collected real-world flawed designs (Figure 5) and mapped their defects to the operations in Algorithm 1. As observed in the "Startup Job Fair" poster, the text blends into the dark background, rendering it illegible. This specific error is simulated in our pipeline by \textbf{Color Perturbation} module. Similarly, the "End Violence" poster exhibits severe layering issues where graphical elements obstruct critical text. This type of flaw is effectively reproduced by our \textbf{Layout Perturbation} module, which applies random coordinate shifts (Position Shift) to elements, creating unintended overlaps and layout collisions. This structural alignment confirms that our synthetic pipeline generates meaningful negatives align with the real world flaw designs.}

\textcolor{black}{\textbf{Note on Dataset Diversity and Benchmark Showcase.} It is important to note that for the Benchmark Showcase (Sec. ~\ref{benchmark showcase}), we intentionally selected examples with isolated and obvious flaws. This selection strategy was adopted strictly for pedagogical purposes—to provide clear, unambiguous visualizations of each specific aesthetic indicator (e.g., illustrating exactly what a "Balance" violation looks like in isolation). Readers should be aware that the full AesEval-Bench and AesEval-Train datasets are significantly more diverse and challenging. They encompass a wide spectrum of difficulty, ranging from the clear, single-flaw examples shown in the showcase to complex, multi-flaw designs (similar to the real-world examples in Fig. ~\ref{fig:real world}) where multiple indicators (e.g., Alignment, Legibility, and Color Harmony) may be compromised simultaneously.}

\textcolor{black}{\section{Task-Specific Prompt Showcases.}}

\textcolor{black}{To ensure a rigorous evaluation of reasoning-augmented models (e.g., GPT-o1, GPT-o3), we also formulated a specific set of Optimized Prompts designed to elicit their chain-of-thought capabilities. As detailed in Tab ~\ref{tab:prompt_optimization}, distinct from the direct queries used for standard VLMs ("Original Prompt"), these optimized prompts explicitly instruct the model to engage in a step-by-step analytical process. However, even task-specifically designed prompts could not improve the model's aesthetic understanding capability, revealing the limitations of general reasoning in this task.}

\begin{table*}[t]
\centering 
\renewcommand{\arraystretch}{1.5}
\setlength{\tabcolsep}{4pt} 

\begin{tabularx}{\linewidth}{ 
    >{\bfseries\color{black}\raggedright}p{2.2cm} 
    >{\color{black}\raggedright\arraybackslash}X 
    >{\color{black}\raggedright\arraybackslash}X 
}
\toprule
Task & Original Prompt & Optimized Prompt \\
\midrule

Aesthetic Judgment & 
\$Indicators\$ + Answer it with one word `yes' or `no'. & 
Analyze the design based on \$Indicators\$. Evaluate the visual elements step-by-step to determine if they meet the standard. Then answer with one word `yes' or `no'. \\
\cmidrule{1-3}

Region Selection & 
\$Indicators\$ + Please only provide the index of Not aesthetic element in given bbox choices. A. [] B.[] C.[] D.[]. & 
Examine the candidate regions (A, B, C, D) regarding \$Indicators\$. Reason through the visual details to identify which specific region exhibits the flaw. Then output the index of that element. \\
\cmidrule{1-3}

Precise Localization & 
\$Indicators\$ + Please only provide bounding box of the Not aesthetic element... If there is no any problems, please return `None'. & 
Analyze the entire design to pinpoint any element that violates \$Indicators\$. If a flaw is found, step-by-step determine its exact spatial coordinates. Output the bounding box in ... format, or return `None'. \\

\bottomrule
\end{tabularx}

\caption{\textcolor{black}{Comparison of Original Prompts vs. Optimized Prompts for Reasoning Models.}}
\label{tab:prompt_optimization}
\vspace{-6mm}
\end{table*}

\textcolor{black}{\section{Statistics tests}}
{
\color{black} 
We conducted rigorous hypothesis testing to ensure the reliability of our results reported in Table 6.

\begin{itemize}
    \item \textbf{Aesthetic Judgment \& Region Selection:} Since these are classification tasks, we applied McNemar's Test to analyze the discordance between models. The results show significant differences, with p-values of $p=0.004$ and $p < 0.001$, respectively, confirming the effectiveness of our fine-tuning pipeline.
    
    \item \textbf{Precise Localization:} For IoU scores, we performed a Paired T-Test. Our fine-tuned model achieves a significant lead over the strongest baseline (GPT-5), with a t-statistic of $t=4.82$ and a p-value of $p < 0.01$.
\end{itemize}
}

\section{Benchmark Showcase}
\label{benchmark showcase}
In this section, we provide visual examples to better illustrate the evaluation criteria for the various aesthetic dimensions within the AesEval-Benchmark. As shown in Table \ref{tab: alignment}, Table \ref{tab:balance}, Table \ref{tab:color}, Table \ref{tab:hierachy}, Table \ref{tab:layering}, Table \ref{tab:legbility} and Table \ref{tab:quality}, each showcase presents a side-by-side comparison of designs that exemplify positive and negative attributes for a specific criterion. These examples serve to clarify the standards used for judging aspects such as layout alignment, color harmony, graphic quality, and font legibility, offering a tangible guide to our benchmark's methodology.

\begin{table*}[ht]

\renewcommand{\arraystretch}{2}

\setlength{\tabcolsep}{0.5mm}

\begin{minipage}{1\textwidth}

\centering


\resizebox{\textwidth}{!}{

\scalebox{0.9}{

\begin{tabular}{l p{5.5cm} p{7.5cm} }

\toprule

\multicolumn{3}{l}{\bf Benchmark Showcase}\\

\midrule

\multicolumn{3}{l}{\colorbox{mypink}{\textit{\textbf{$\triangleright$ Layout-Alignment, Graphic-Quality}}}} \\

&\multicolumn{1}{c}{\includegraphics[height=3cm]{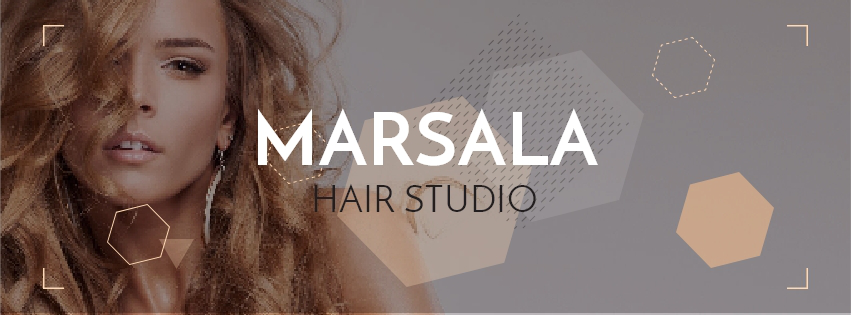}}

&\multicolumn{1}{c}{\includegraphics[height=3cm]{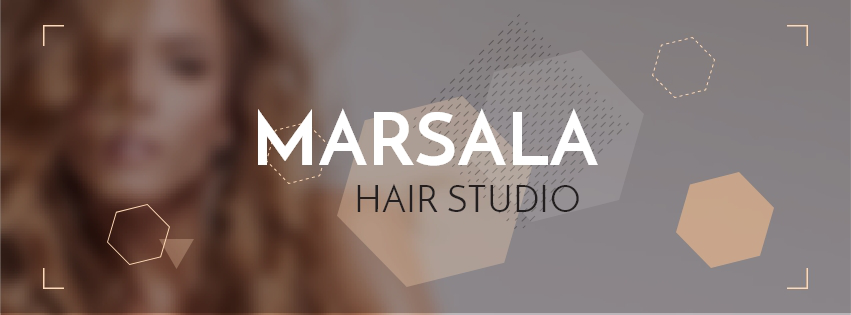}} \\

\midrule

Explanation:

& Clear background, center-aligned text.

& The background is blurry and the words in the middle are not aligned.

\\

\bottomrule

\end{tabular}

}

}

\caption{Examples of Layout-Alignment and Graphic-Quality in AesEval-Benchmark.}

\label{tab: alignment}
\end{minipage}
\end{table*}

\begin{table*}[ht]
 \renewcommand{\arraystretch}{2}
 \setlength{\tabcolsep}{0.5mm}
   \begin{minipage}{1\textwidth}
 \centering 
 \scalebox{0.9}{
 \begin{tabular}{l p{5.5cm} p{7.5cm} }
 \toprule
  \multicolumn{3}{l}{\bf Benchmark Showcase}   

  \\
 \midrule
 \multicolumn{3}{l}{\colorbox{mypink}{\textit{\textbf{$\triangleright$ Font-Legbility}}}} \\
 &  \multicolumn{1}{c}{\includegraphics[height=3cm]{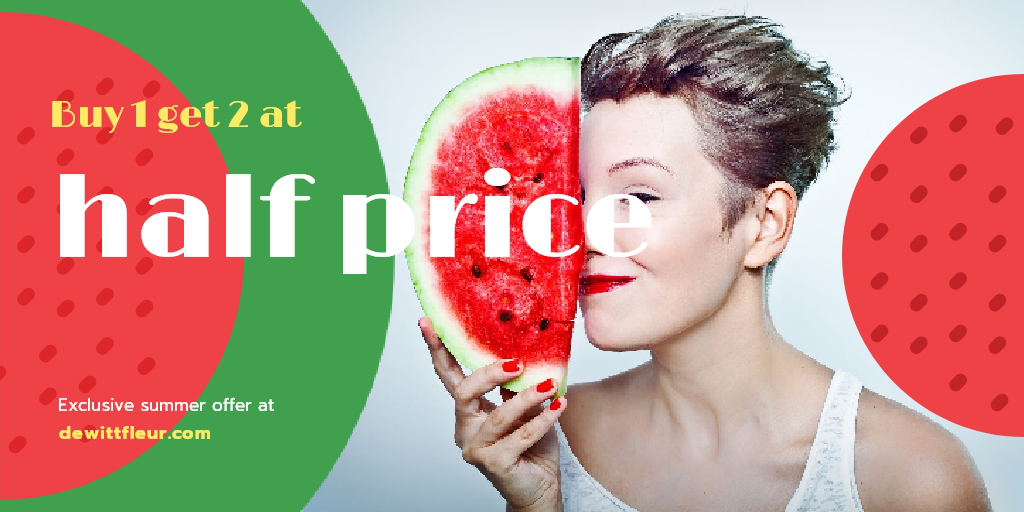}}
 &  \multicolumn{1}{c}{\includegraphics[height=3cm]{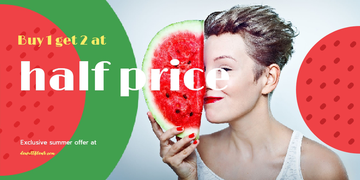}} \\
 \midrule
 Explanation:
 & The yellow font in the lower left corner is clearly visible.
 & The yellow text in the lower left corner becomes blurred and invisible.
 \\
 \bottomrule
 \end{tabular}
 }
 \caption{Examples of Font-Legbility in AesEval-Benchmark.}
 \label{tab:legbility}   \end{minipage}
 \end{table*}

\begin{table*}[ht]
 \renewcommand{\arraystretch}{2}
 \setlength{\tabcolsep}{0.5mm}
   \begin{minipage}{1\textwidth}
 \centering 
 \scalebox{0.9}{
 \begin{tabular}{l p{5.5cm} p{7.5cm} }
 \toprule
  \multicolumn{3}{l}{\bf Benchmark Showcase}   

  \\
 \midrule
 \multicolumn{3}{l}{\colorbox{mypink}{\textit{\textbf{$\triangleright$ Graphic-Relevance}}}} \\
 &  \multicolumn{1}{c}{\includegraphics[height=3cm]{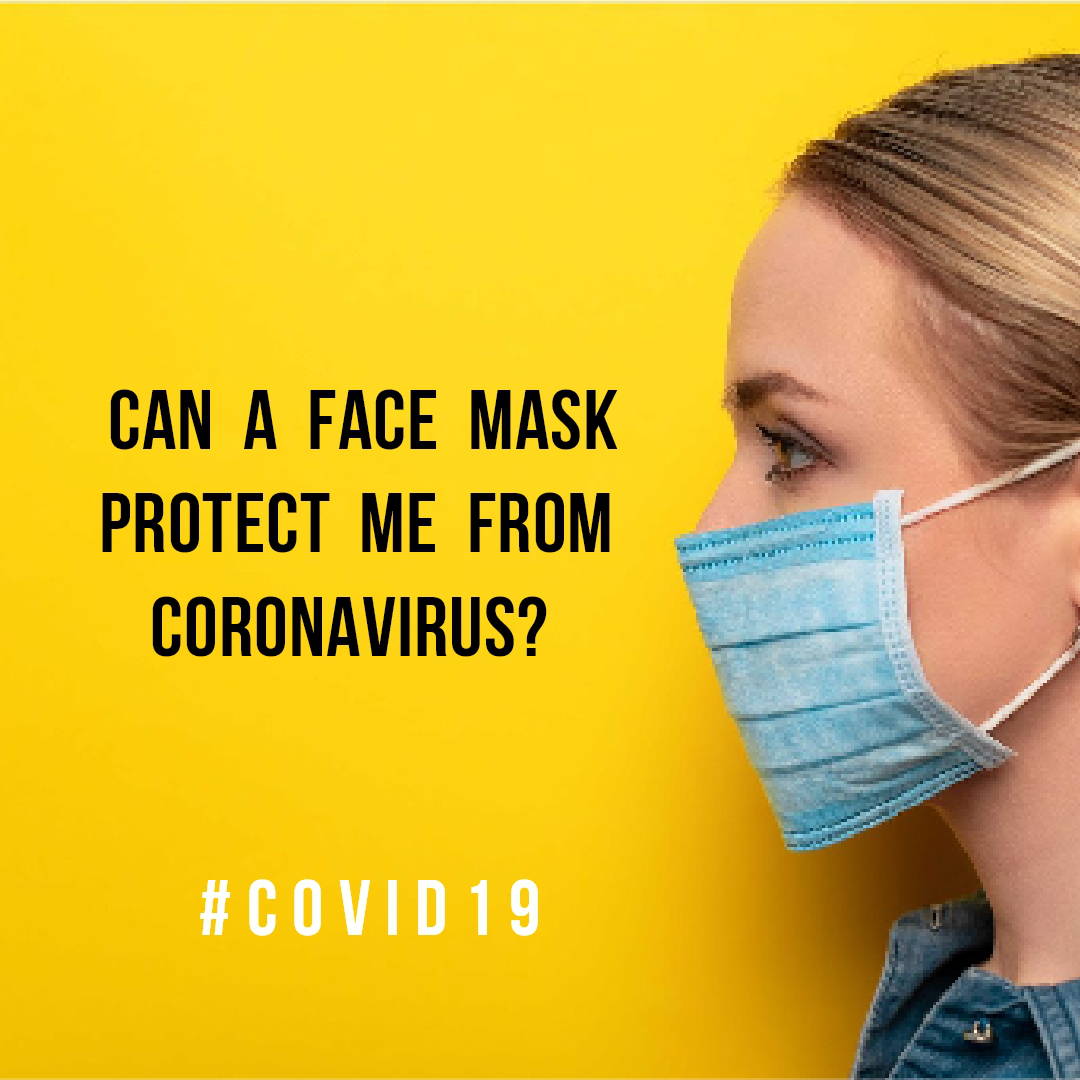}}
 &  \multicolumn{1}{c}{\includegraphics[height=3cm]{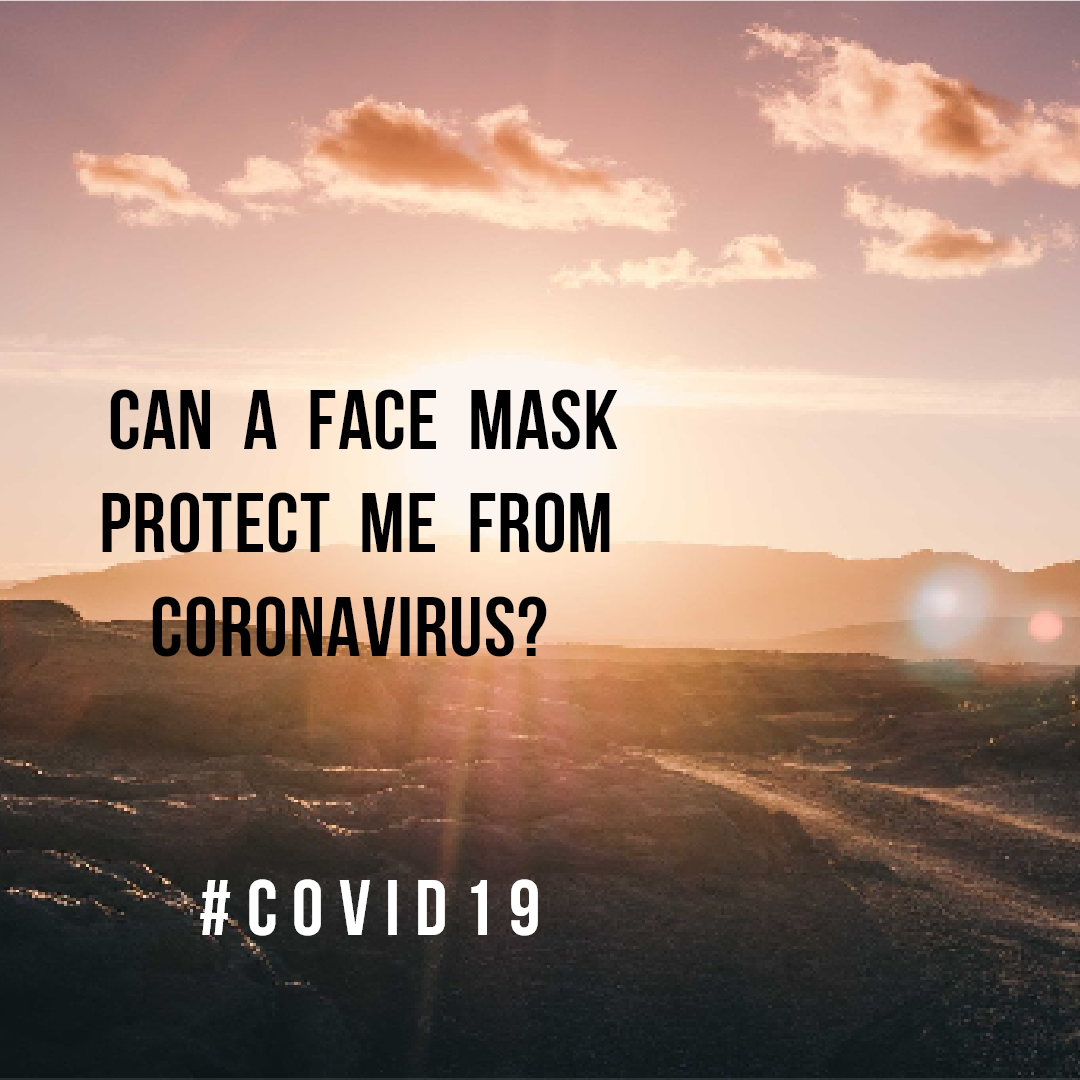}} \\
 \midrule
 Explanation:
 & The background is a woman wearing a mask, which is relevant.
 & The background is a beautiful landscape photo, which does not fit the theme.
 \\
 \bottomrule
 \end{tabular}
 }
 \caption{Examples of Graphic-Relevance in AesEval-Benchmark.}
 \label{tab:legbility}   \end{minipage}
 \end{table*}

\begin{table*}[ht]
 \renewcommand{\arraystretch}{2}
 \setlength{\tabcolsep}{0.5mm}
   \begin{minipage}{1\textwidth}
 \centering 
 \scalebox{0.9}{
 \begin{tabular}{l p{5.5cm} p{7.5cm} }
 \toprule
  \multicolumn{3}{l}{\bf Benchmark Showcase}   

  \\
 \midrule
 \multicolumn{3}{l}{\colorbox{mypink}{\textit{\textbf{$\triangleright$ Layout-Whitespace, Layout-Layering}}}} \\
 &  \multicolumn{1}{c}{\includegraphics[height=3cm]{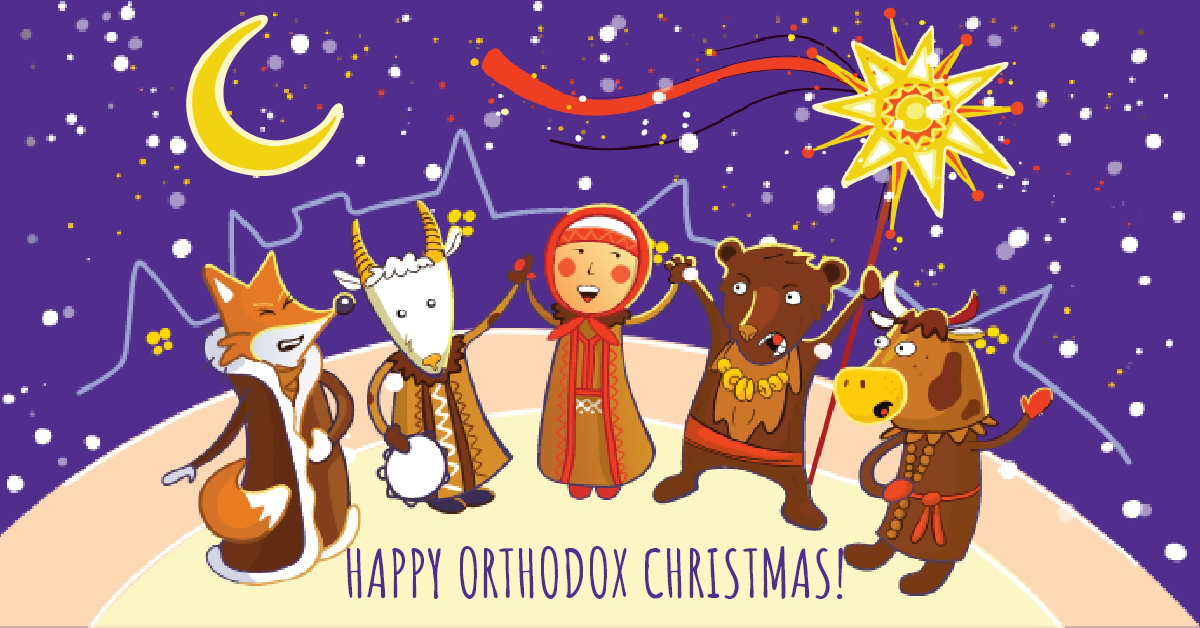}}
 &  \multicolumn{1}{c}{\includegraphics[height=3cm]{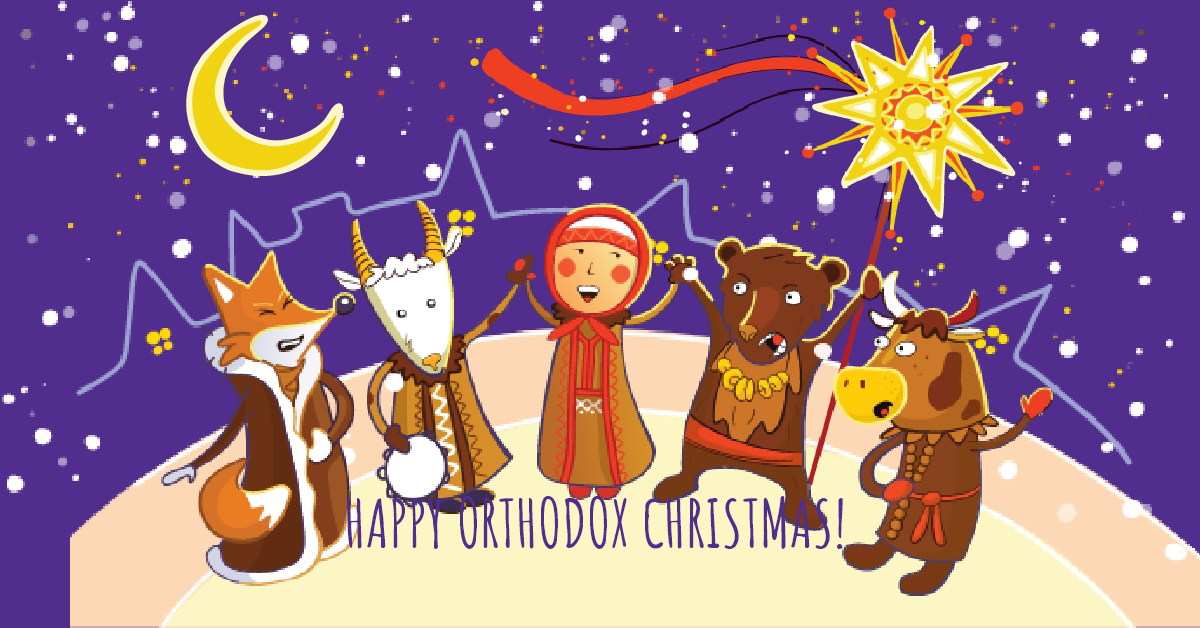}} \\
 \midrule
 Explanation:
 & The text is in the blank space , with no blank space or overlap.
 & The text has been moved to the animal's feet, resulting in white space below and stacked elements.
 \\
 \bottomrule
 \end{tabular}
 }
 \caption{Examples of Layout-Whitespace and Layout-Layering in AesEval-Benchmark.}
 \label{tab:layering}   \end{minipage}
 \end{table*}

\begin{table*}[ht]
 \renewcommand{\arraystretch}{2}
 \setlength{\tabcolsep}{0.5mm}
   \begin{minipage}{1\textwidth}
 \centering 
 \scalebox{0.9}{
 \begin{tabular}{l p{5.5cm} p{7.5cm} }
 \toprule
  \multicolumn{3}{l}{\bf Benchmark Showcase}   

  \\
 \midrule
 \multicolumn{3}{l}{\colorbox{mypink}{\textit{\textbf{$\triangleright$ Layout-Balance}}}} \\
 &  \multicolumn{1}{c}{\includegraphics[height=3cm]{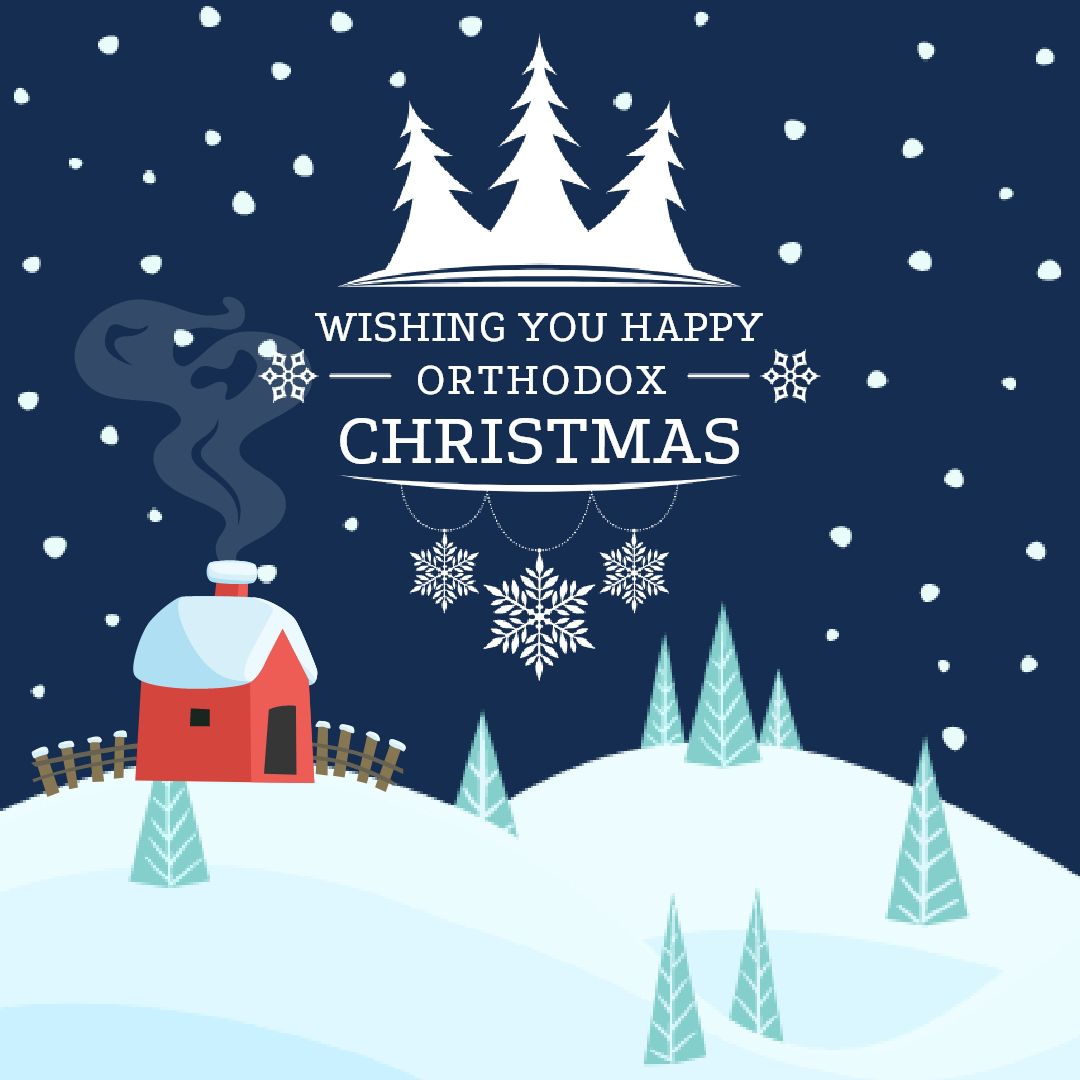}}
 &  \multicolumn{1}{c}{\includegraphics[height=3cm]{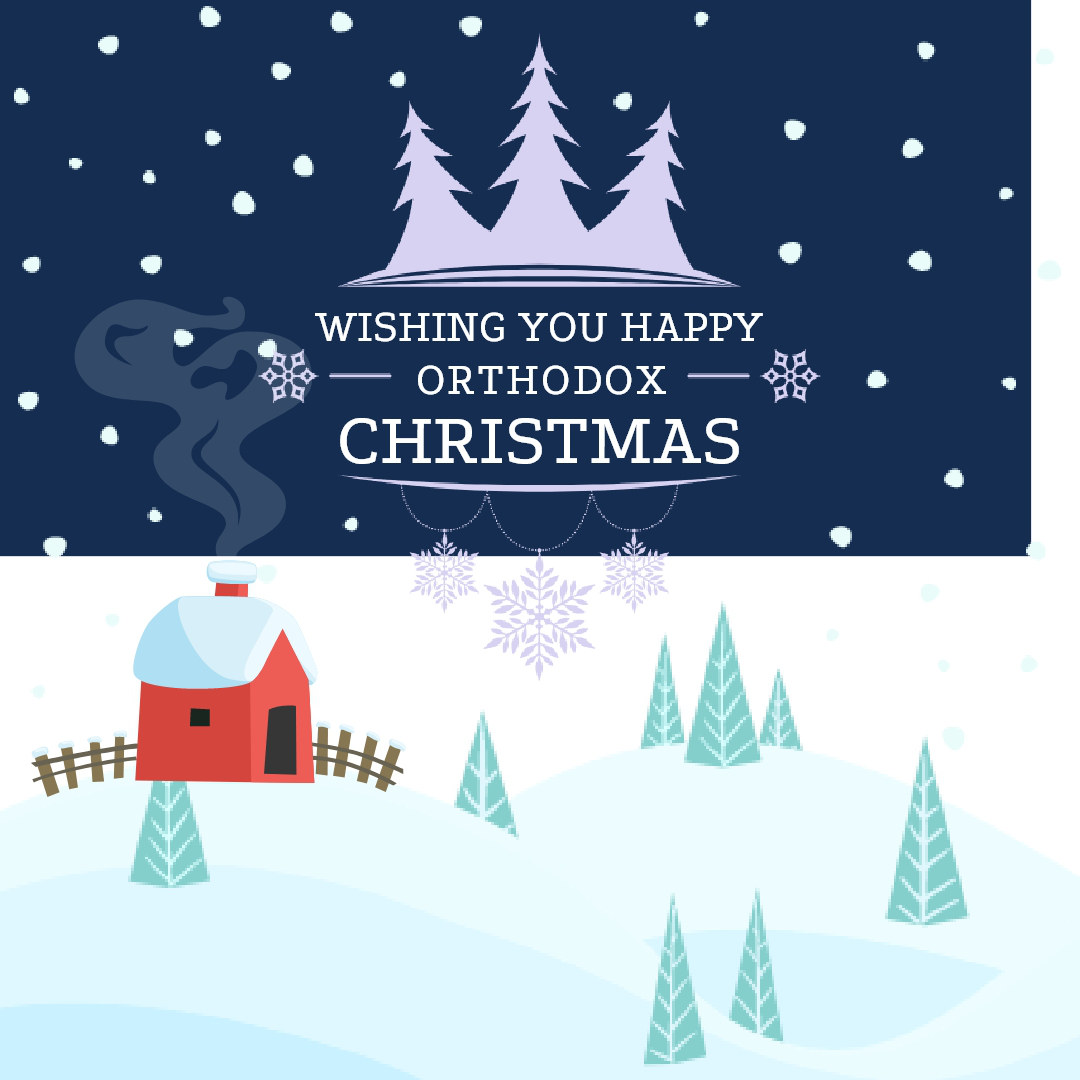}} \\
 \midrule

 Explanation:
 & The overall layout of the picture is balanced and coordinated.
 & The background of the picture is moved upwards, and the balance of the whole picture is broken.
 \\
 \bottomrule
 \end{tabular}
 }
 \caption{Examples of Layout-Balance in AesEval-Benchmark.}
 \label{tab:balance}   \end{minipage}
 \end{table*}

\begin{table*}[ht]
 \renewcommand{\arraystretch}{2}
 \setlength{\tabcolsep}{0.5mm}
   \begin{minipage}{1\textwidth}
 \centering 
 \scalebox{0.9}{
 \begin{tabular}{l p{5.5cm} p{7.5cm} }
 \toprule
  \multicolumn{3}{l}{\bf Benchmark Showcase}   

  \\
 \midrule
 \multicolumn{3}{l}{\colorbox{mypink}{\textit{\textbf{$\triangleright$ Layout-Hierarchy}}}} \\
 &  \multicolumn{1}{c}{\includegraphics[height=3cm]{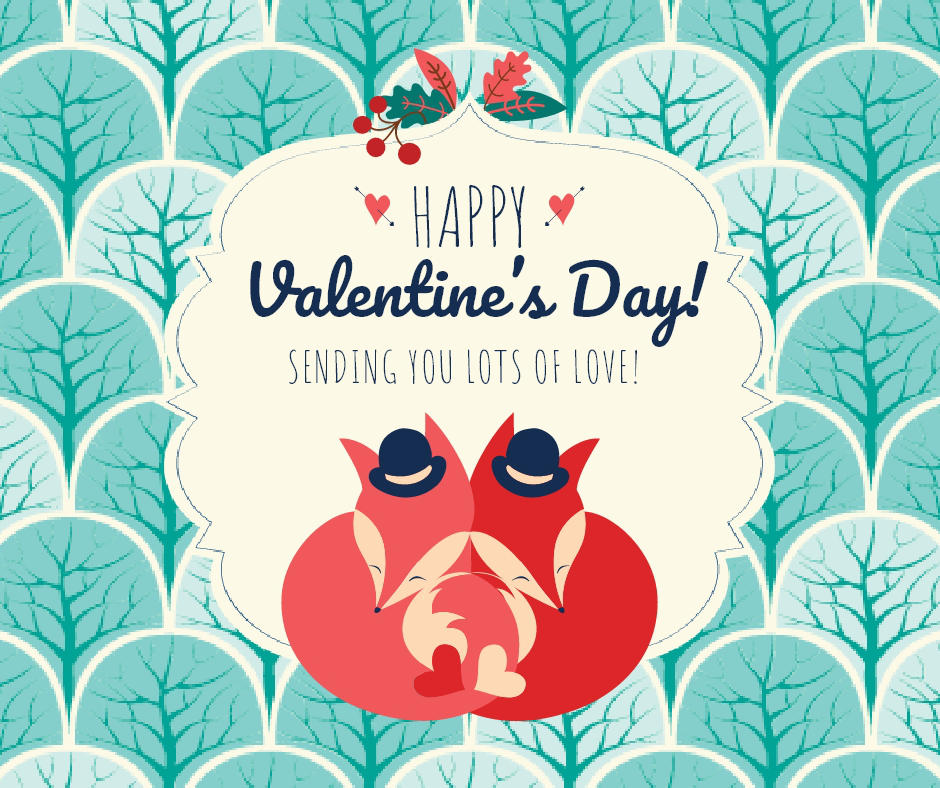}}
 &  \multicolumn{1}{c}{\includegraphics[height=3cm]{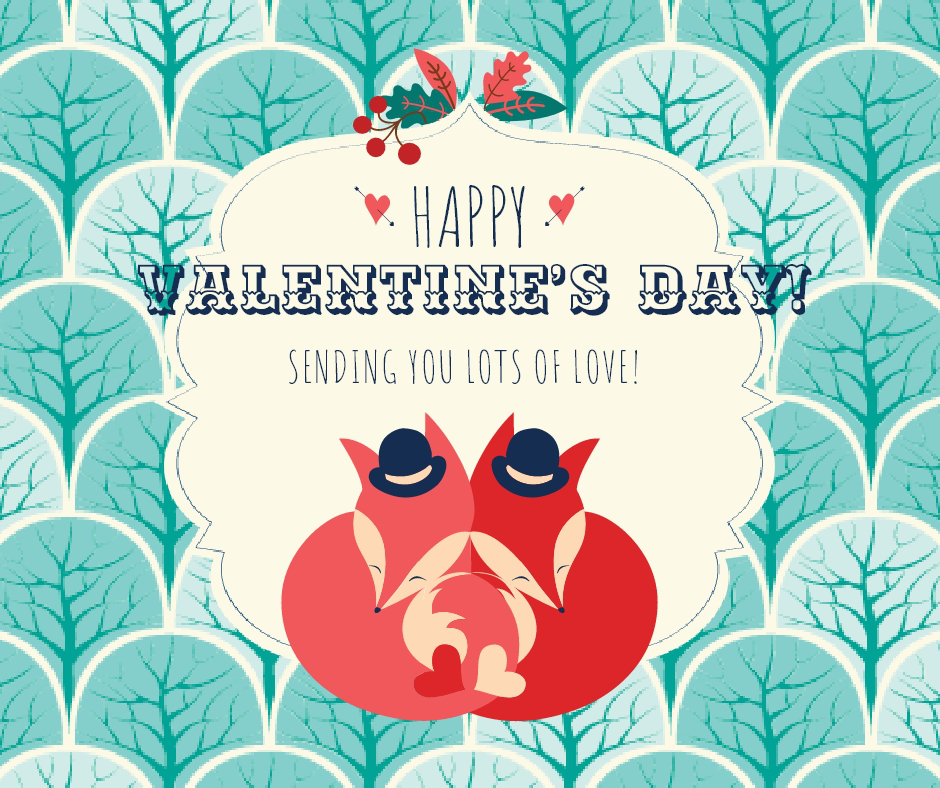}} \\
 \midrule
 Explanation:
 & All the fonts in the picture are consistent, which gives a sense of hierarchy.
 & The font in the picture is disturbed and looks like it is not on the same level as the previous font. There is no sense of hierarchy.
 \\
 \bottomrule
 \end{tabular}
 }
 \caption{Examples of Layout-Hierarchy in AesEval-Benchmark.}
 \label{tab:hierachy}   \end{minipage}
 \end{table*}

\begin{table*}[ht]
 \renewcommand{\arraystretch}{2}
 \setlength{\tabcolsep}{0.5mm}
   \begin{minipage}{1\textwidth}
 \centering 
 \resizebox{\textwidth}{!}{
 \scalebox{0.9}{
 \begin{tabular}{l p{5.5cm} p{7.5cm} }
 \toprule
  \multicolumn{3}{l}{\bf Benchmark Showcase}   

  \\
 \midrule
 \multicolumn{3}{l}{\colorbox{mypink}{\textit{\textbf{$\triangleright$ Graphic-Quality, Color-Contrast}}}} \\
 &  \multicolumn{1}{c}{\includegraphics[height=3cm]{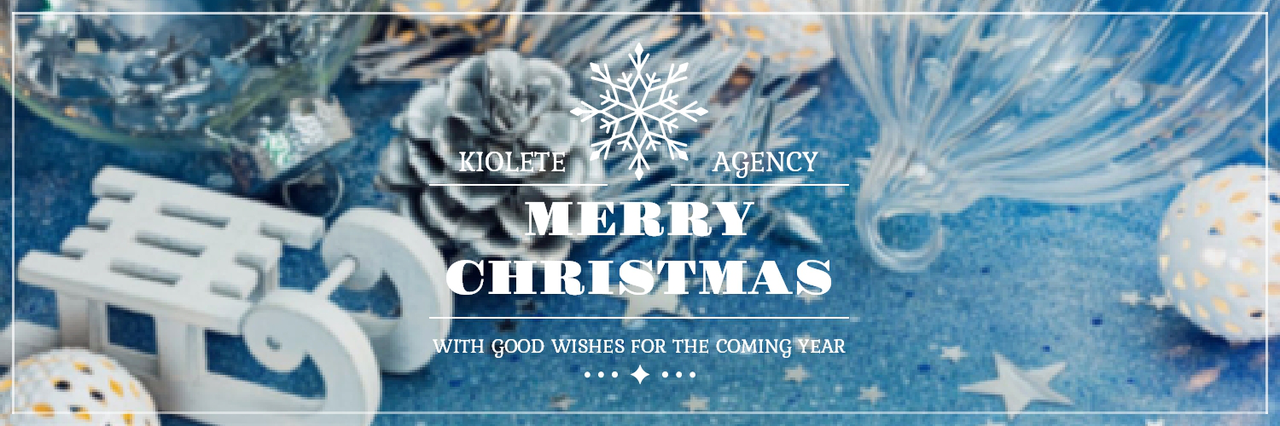}}
 &  \multicolumn{1}{c}{\includegraphics[height=3cm]{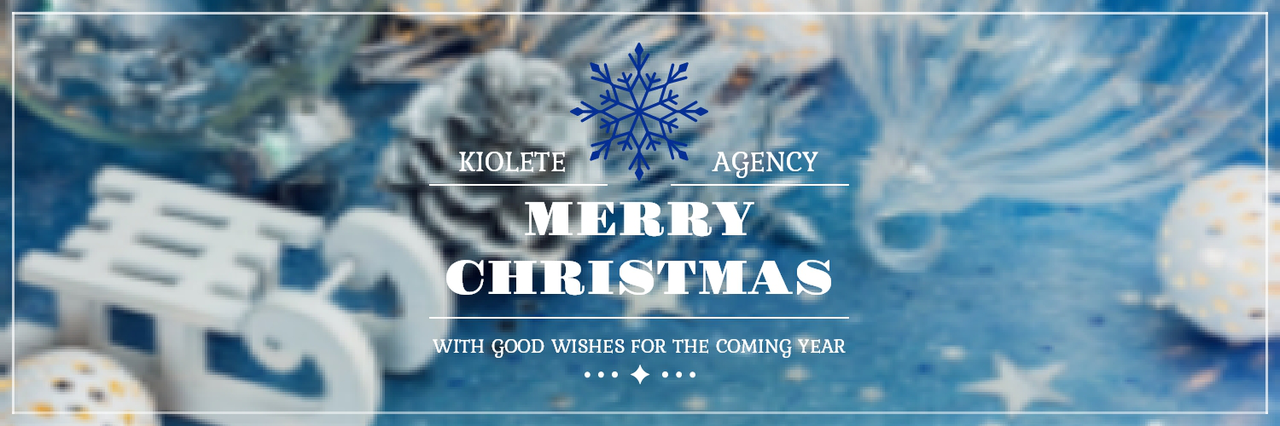}} \\
 \midrule

 Explanation:
 & Clear background, black and white colors have good contrast.
 & The background is blurred and the color changes from white to black, with no contrast to the background.
 \\
 \bottomrule
 \end{tabular}
 }}
 \caption{Examples of Graphic-Quality and Color-Contrast in AesEval-Benchmark.}
 \label{tab:quality}   \end{minipage}
 \end{table*} 

\begin{table*}[ht]
 \renewcommand{\arraystretch}{2}
 \setlength{\tabcolsep}{0.5mm}
   \begin{minipage}{1\textwidth}
 \centering 
 \scalebox{0.9}{
 \begin{tabular}{l p{5.5cm} p{7.5cm} }
 \toprule
  \multicolumn{3}{l}{\bf Benchmark Showcase}   

  \\
 \midrule
 \multicolumn{3}{l}{\colorbox{mypink}{\textit{\textbf{$\triangleright$ Color-Harmony, Color-Appealing, Color-Psychology}}}} \\
 &  \multicolumn{1}{c}{\includegraphics[height=3cm]{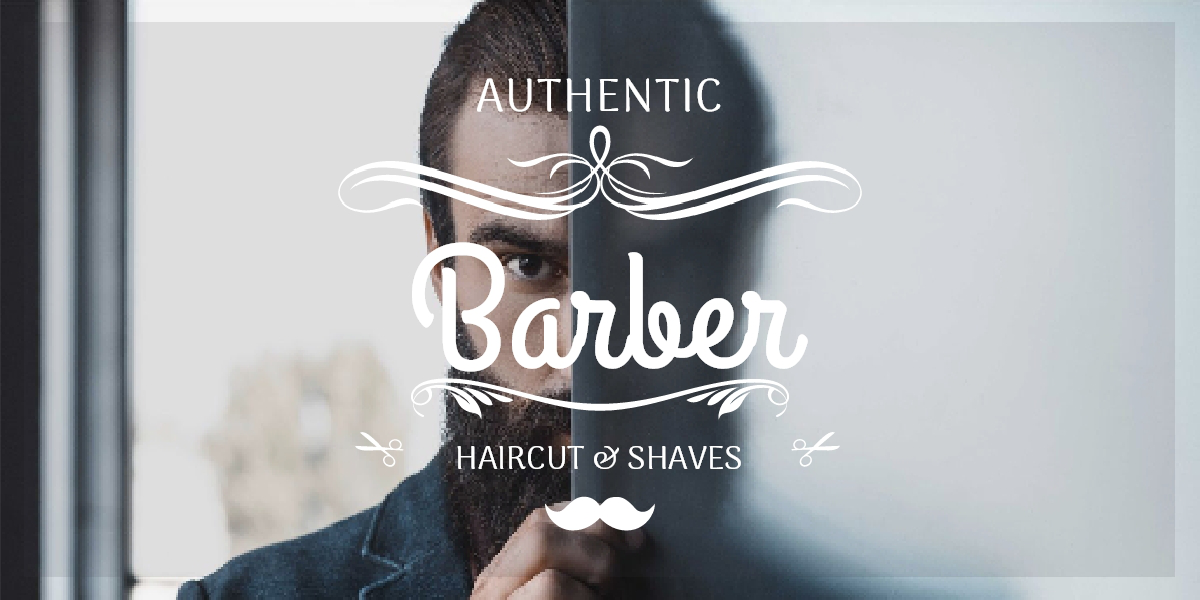}}
 &  \multicolumn{1}{c}{\includegraphics[height=3cm]{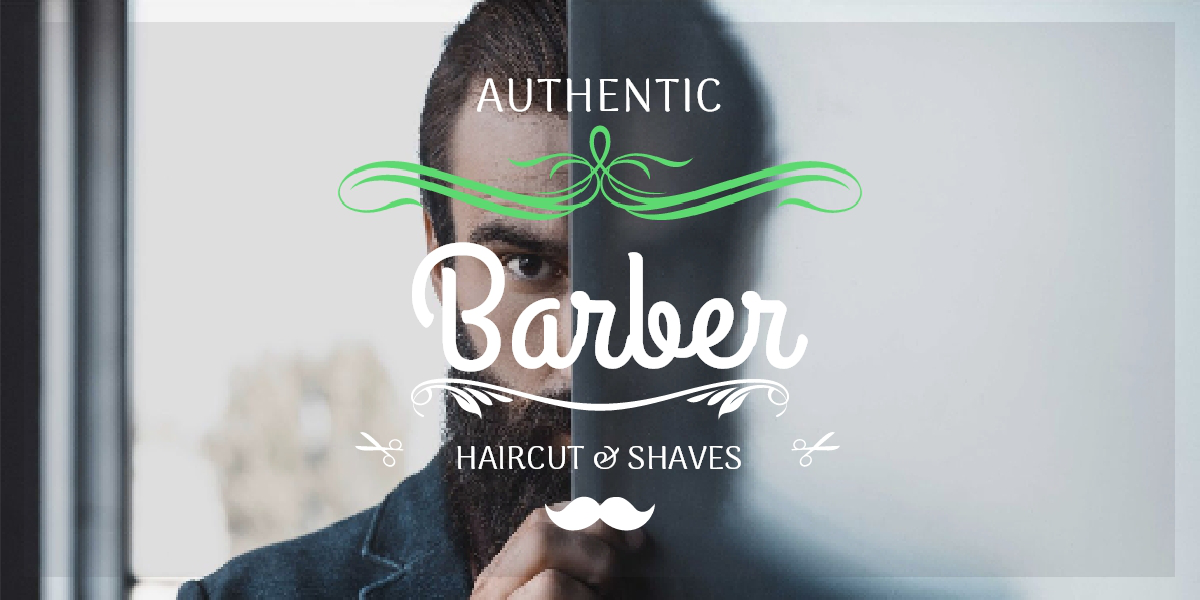}} \\
 \midrule

 Explanation:
 & The whole picture has harmonious and beautiful colors.
 & The middle element turns green, which makes the whole picture look disharmonious and unappealing. Green is strange and cause bad psychological effects.
 \\
 \bottomrule
 \end{tabular}
 }
 \caption{Examples of Color-Harmony, Color-Appealing and Color-Psychology in AesEval-Benchmark.}
 \label{tab:color}   \end{minipage}
 \end{table*}

\end{document}